\documentclass[10pt,twocolumn,letterpaper]{article}
\usepackage[pagenumbers]{cvpr}
\usepackage[T1]{fontenc}
\usepackage{mathtools}
\usepackage{microtype}
\usepackage{enumitem}
\usepackage{algorithm}
\usepackage{algpseudocode}
\usepackage{multirow}
\newcommand{\method}{SAMR\xspace}
\newcommand{\R}{\mathbb{R}}
\newcommand{\E}{\mathbb{E}}
\newcommand{\norm}[1]{\left\lVert #1\right\rVert}
\DeclareMathOperator*{\argmin}{arg\,min}
\newlength{\paraheadskip}
\renewcommand{\paragraph}[1]{\vspace{\paraheadskip}\noindent\textbf{#1}\hspace{0.5em}}
\newcommand{\llcal}{LongLive calib.\xspace}
\newcommand{\MainLatency}{27.47}
\newcommand{\ReuseLatency}{30.13}
\newcommand{\LegacyLatency}{29.43}
\newcommand{\FullLatency}{34.71}
\newcommand{\MainPSNR}{15.47}
\newcommand{\ReusePSNR}{13.87}
\newcommand{\LegacyPSNR}{14.92}
\newcommand{\PSNRGain}{1.61}

\newcommand{\FullLatencyReduction}{20.9}
\newcommand{\FullSpeedup}{1.26}
\newcommand{\MainTFLOPs}{2799.5}
\newcommand{\ReuseTFLOPs}{3242.3}
\newcommand{\FullTFLOPs}{4041.4}
\newcommand{\FullTFLOPCut}{30.7}

\newcommand{\DenoiseTFLOPSaved}{446.9}

\newcommand{\MainReconTFLOPs}{3.09}
\newcommand{\ReconFLOPRatio}{144}
\newcommand{\FullCommitStepOneS}{20.57}
\newcommand{\FullDenoiseStageS}{13.96}
\newcommand{\ReuseCommitStepOneS}{14.98}
\newcommand{\ReuseDenoiseStageS}{14.94}

\newcommand{\MainCommitStepOneS}{15.54}
\newcommand{\MainDenoiseStageS}{11.83}
\newcommand{\RoutingSpeedup}{8.47}
\newcommand{\RoutingOld}{764.9}
\newcommand{\RoutingNew}{90.3}
\newcommand{\MainSteps}{1.58}
\newcommand{\ReuseSteps}{1.87}
\newcommand{\LegacySteps}{1.72}
\newcommand{\FullSteps}{1.78}
\newcommand{\MainRefresh}{24.7}
\newcommand{\ReuseRefresh}{24.6}
\newcommand{\MainReconPct}{45.2}
\newcommand{\CostModelMAE}{5.07}
\newcommand{\CostModelRelErr}{2.1}
\newcommand{\CorrSpeedup}{27.6}
\newcommand{\PlanSpeedup}{11.7}
\newcommand{\MainDitDelta}{-2.66}
\newcommand{\MainDitCILo}{-4.07}
\newcommand{\MainDitCIHi}{-1.45}
\newcommand{\MainPSNRCILo}{+1.12}
\newcommand{\MainPSNRCIHi}{+2.12}
\newcommand{\MainSatDelta}{-2.47}
\newcommand{\MainSatCILo}{-5.67}
\newcommand{\MainSatCIHi}{+0.71}
\newcommand{\MainFlkDelta}{-0.36}
\newcommand{\MainFlkCILo}{-6.38}
\newcommand{\MainFlkCIHi}{+4.99}
\newcommand{\MainMotDelta}{-0.22}
\newcommand{\MainMotCILo}{-6.94}
\newcommand{\MainMotCIHi}{+5.57}
\newcommand{\LegacySat}{7.99}
\newcommand{\LegacyFlk}{20.63}
\newcommand{\LegacyMot}{18.73}

\newcommand{\MainFlk}{23.46}
\newcommand{\MainMot}{22.81}
\newcommand{\UntouchedDit}{30.41}

\newcommand{\UntouchedFullDit}{36.71}
\newcommand{\UntouchedSpeedup}{1.21}
\newcommand{\UntouchedSteps}{1.82}
\newcommand{\UntouchedReuseSteps}{2.12}
\newcommand{\UntouchedFullSteps}{1.97}

\newcommand{\UntouchedTFLOPCut}{27.3}

\newcommand{\UntouchedGain}{1.14}
\newcommand{\UntouchedDitDelta}{-2.33}
\newcommand{\UntouchedPSNRWins}{15}
\newcommand{\UntouchedSpeedWins}{12}
\newcommand{\DelivFullSerial}{70.22}

\newcommand{\DelivAsyncVsFullGPUCut}{16.0}
\newcommand{\DelivSerial}{63.23}
\newcommand{\DelivAsync}{58.98}
\newcommand{\DelivTwoGPU}{37.79}

\newcommand{\DelivAsyncFPS}{13.03}
\newcommand{\DelivTwoGPUFPS}{20.25}
\newcommand{\DelivSerialFirst}{0.903}
\newcommand{\DelivAsyncFirst}{1.256}

\newcommand{\DelivAsyncCut}{6.7}

\newcommand{\DelivAsyncGPU}{59.0}
\newcommand{\DelivTwoGPUSec}{75.6}
\newcommand{\ScreenDit}{30.11}
\newcommand{\ScreenPSNR}{15.55}
\newcommand{\AbsFullVB}{93.56}
\newcommand{\AbsReuseVB}{92.10}

\newcommand{\AbsOursVB}{92.77}

\newcommand{\AbsCFVB}{93.41}

\newcommand{\AbsReuseSSIM}{0.529}

\newcommand{\AbsOursSSIM}{0.552}

\newcommand{\AbsReuseLPIPS}{0.831}

\newcommand{\AbsOursLPIPS}{0.780}

\newcommand{\AbsOursSmooth}{97.40}
\newcommand{\AbsReuseSmooth}{96.70}

\newcommand{\AbsCFDit}{32.85}

\newcommand{\SFDFullDit}{11.73}
\newcommand{\SFDReuseDit}{10.65}
\newcommand{\SFDOursDit}{10.43}

\newcommand{\SFDCFDynDit}{9.69}
\newcommand{\SFDFullSteps}{2.22}
\newcommand{\SFDReuseSteps}{2.47}
\newcommand{\SFDOursSteps}{2.30}

\newcommand{\SFDOursPSNR}{14.61}
\newcommand{\SFDReusePSNR}{14.28}
\newcommand{\SFDPSNRGain}{0.33}
\newcommand{\SFDPSNRWins}{13}

\newcommand{\SFDOursLPIPS}{0.820}
\newcommand{\SFDReuseLPIPS}{0.817}

\newcommand{\SFDIndomainPSNR}{14.77}
\newcommand{\SFDIndomainVB}{91.10}
\newcommand{\SFDIndomainLPIPS}{0.798}
\newcommand{\SFDFullVB}{92.58}
\newcommand{\SFDZeroshotVB}{89.62}
\newcommand{\SFDReuseVB}{91.16}

\newcommand{\SFDCFDynVB}{92.10}
\newcommand{\SFDCFDynPSNR}{16.32}
\newcommand{\SFDCFDynLPIPS}{0.654}

\newcommand{\SFDNoCommitLPIPS}{0.881}
\newcommand{\WaterfallBaseLong}{65.30}
\newcommand{\WaterfallFastLong}{25.44}
\newcommand{\WaterfallSpdLong}{2.57}
\newcommand{\WaterfallSpdWide}{2.28}
\newcommand{\WaterfallSpdSFD}{2.13}
\newcommand{\FloorEpsSmallPSNR}{22.60}
\newcommand{\FloorEpsLargePSNR}{21.91}

\definecolor{cvprblue}{rgb}{0.21,0.49,0.74}
\usepackage[pagebackref,breaklinks,colorlinks,allcolors=cvprblue]{hyperref}
\usepackage{colortbl}
\hypersetup{pdftitle={Cheap Memory Costs Future Compute: Co-Designing Persistent KV and GPU Execution for Efficient Streaming Video Diffusion},pdfauthor={Jian Yang, Zhizhuo KOU, Qiyuan Zhu, CHEN Han, Sirui Han, Yike Guo}}
\def\confName{CVPR}
\def\confYear{2027}
\title{Cheap Memory Costs Future Compute: Co-Designing Persistent KV and\\GPU Execution for Efficient Streaming Video Diffusion}
\author{%
\normalsize Jian Yang$^{1}$, Zhizhuo KOU$^{1}$, Qiyuan Zhu$^{1}$, CHEN Han$^{2}$, Sirui Han$^{1}$, Yike Guo$^{1}$\\[4pt]
\small $^{1}$ The Hong Kong University of Science and Technology (ust.hk)\\
\small $^{2}$ National University of Singapore (nus.edu)
}
\begin{document}
\twocolumn[{%
\renewcommand\twocolumn[1][]{#1}%
\maketitle
\begin{center}
\vspace{-1.2em}
\includegraphics[width=\textwidth]{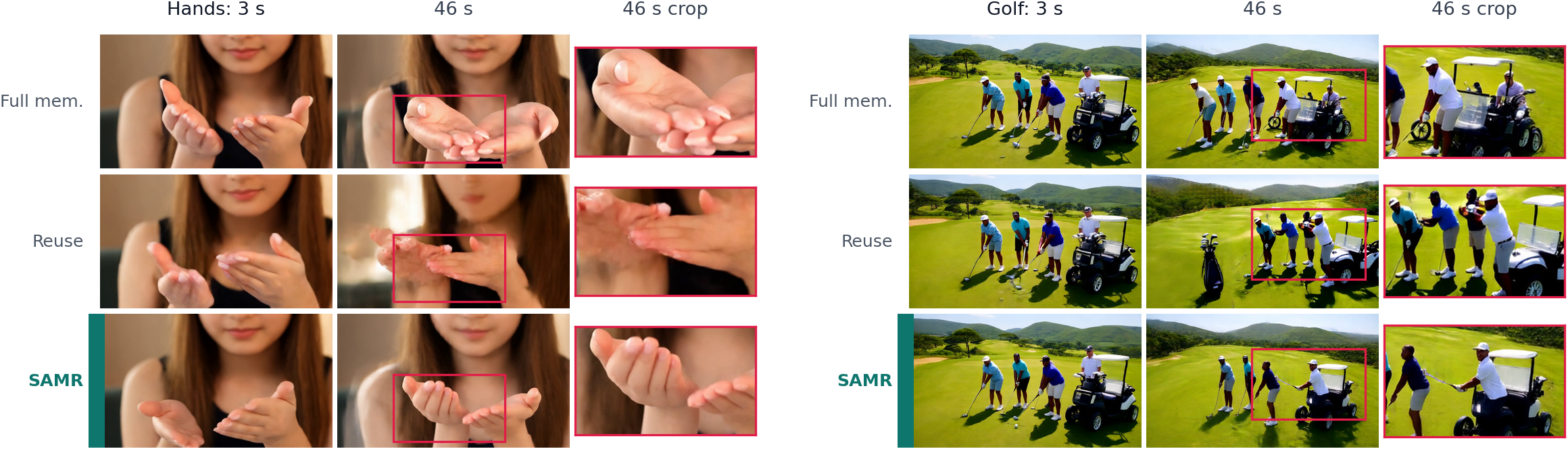}
\vspace{-1.4em}
\captionof{figure}{\textbf{Cheap memory costs late-horizon quality.} Same-seed 48-second streams from frozen LongLive-1.3B (seed 0; frames at $3$\,s and $46$\,s, plus a $46$\,s zoom). \emph{Full memory} recomputes every clean KV entry; \emph{Eventful reuse} refreshes $24.6\%$ of tokens and keeps stale KV elsewhere, which leaves late-horizon ghosting: distorted fingers in \emph{Hands} ($14.02$\,dB final-quarter PSNR) and duplicated silhouettes in \emph{Golf} ($9.28$\,dB). \method reconstructs the unrefreshed KV instead, preserving anatomical structure ($17.22$\,dB) and foreground coherence ($12.75$\,dB) while running $3.30$\,s faster than reuse and using \FullTFLOPCut{}\% fewer FLOPs than \emph{Full memory} (more in Appendix~\ref{sec:qualitative-extra}).}
\label{fig:qualitative}
\end{center}%
\vspace{1.0em}%
}]
\begin{abstract}
Block-causal video diffusion generates long videos block by block, conditioned on a persistent key--value (KV) cache. After each block is denoised, an extra noise-free ($t=0$) forward pass writes its clean KV into the cache; once distillation and speculative early exit make denoising cheap, this \emph{clean KV update} dominates the runtime. Reusing stale KV shortens the update, but stale memory corrupts the conditioning of later blocks and inflates their denoising steps, so the saved compute reappears downstream. We present \textbf{\method} (\emph{Step-Aware Memory Reconstruction}), a training-free method for frozen causal video DiTs that refreshes only part of the cache and reconstructs the rest without DiT forwards. Our key observation is that, once rotary position phases are removed, motion-aligned historical KV and the in-flight KV a token holds at its exit step are complementary estimates of its clean KV. \method debiases in-flight KV with offline-calibrated per-head affine maps in Pre-RoPE space, fuses the two estimates, and reserves true $t=0$ refreshes for tokens flagged by an attention-readout error proxy, all executed by tile-aligned, synchronization-free Triton kernels. On 48-second LongLive-1.3B videos, \method reduces DiT latency by \FullLatencyReduction{}\% and principal FLOPs by \FullTFLOPCut{}\% relative to dense KV updates, lowers downstream denoising steps, and improves final-quarter PSNR over token reuse by \PSNRGain{}\,dB (\UntouchedGain{}\,dB on held-out VidProM-16). The reconstruction also transfers to frozen Self-Forcing-DMD-1.3B: zero-shot, it reduces reuse-induced step inflation ($\SFDReuseSteps{} \to \SFDOursSteps{}$ steps/token) and improves late PSNR by $+\SFDPSNRGain{}$\,dB, and a one-time training-free calibration recovers the diagnostic score that zero-shot transfer lowers ($\SFDZeroshotVB{} \to \SFDIndomainVB{}$).
\end{abstract}

\section{Introduction}
\label{sec:introduction}

\begin{figure*}[!t]
\centering
\includegraphics[width=0.98\textwidth]{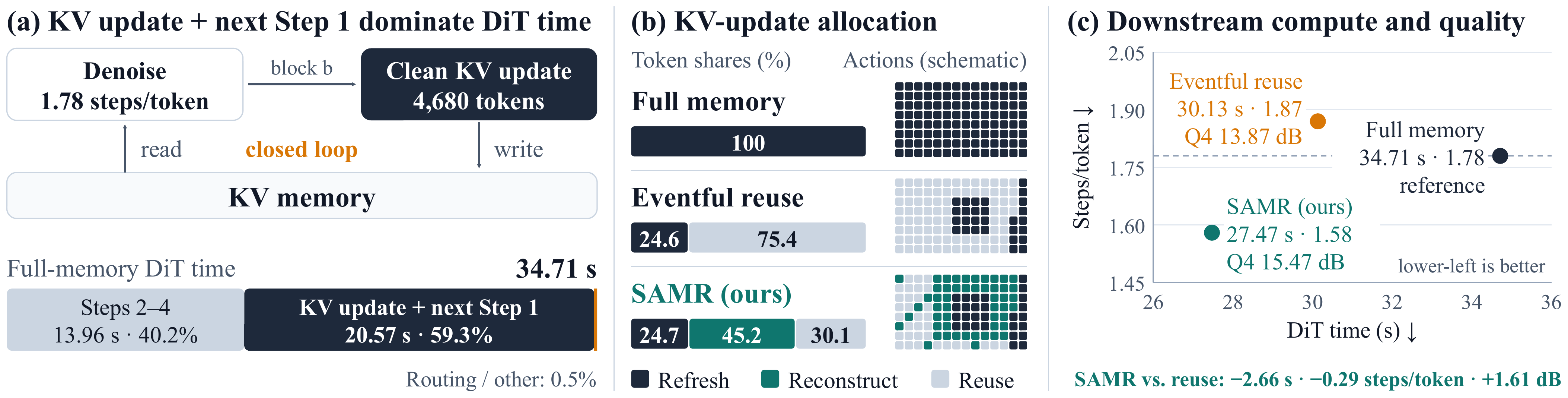}
\caption{\textbf{The clean KV update becomes the bottleneck, and reuse shifts its cost downstream.} \textbf{(a)} Once 4-step distillation and early exit compress denoising to \FullSteps{} steps/token, the merged $t=0$ KV update + next-block Step-1 stage takes $59.3\%$ (\FullCommitStepOneS{}\,s) of the \FullLatency{}\,s full-memory DiT runtime. Denoising reads the cache that this update writes, which closes the loop. \textbf{(b)} At a similar refresh ratio (\ReuseRefresh{}\% vs.\ \MainRefresh{}\%), \emph{Eventful reuse} copies stale KV for $75.4\%$ of tokens, whereas \textbf{\method} reconstructs \MainReconPct{}\% of them in Pre-RoPE space. \textbf{(c)} Reuse pushes downstream denoising above \emph{Full memory} ($\FullSteps{} \to \ReuseSteps{}$ steps/token) at \ReusePSNR{}\,dB final-quarter (Q4) PSNR; \method lowers steps to \MainSteps{} and DiT time to \MainLatency{}\,s ($-\FullLatencyReduction{}\%$ vs.\ dense) while improving Q4 PSNR over reuse by $+\PSNRGain{}$\,dB.}
\label{fig:teaser}
\end{figure*}

Block-causal diffusion turns long-horizon video generation into an autoregressive stream~\cite{yin2025causvid,huang2025selfforcing,yang2025longlive,wan2025} that synthesizes compact frame blocks, each conditioned on a bounded persistent key--value (KV) cache~\cite{xiao2024streamingllm,yi2026deepforcing,jing2026commit}. Every later block reads this cache, so its fidelity determines what the stream looks like tens of seconds later (Fig.~\ref{fig:qualitative}). Because denoising runs on noisy latents at heterogeneous early-exit timesteps, each finished block requires an extra noise-free ($t=0$) forward pass that writes \emph{clean} KV into the cache before later blocks attend to it. As distillation~\cite{yin2024dmd2}, speculative early exit with a merged KV-update and Step-1 pass~\cite{choi2026speculative}, and read-side cache pruning~\cite{ji2026forcingkv,jing2026commit} have compressed per-block denoising, this write-side clean KV update has become the primary bottleneck (Fig.~\ref{fig:teaser}a).

Can the update be made cheaper without retraining? TriTD~\cite{li2026tritd} and FlashForward~\cite{wang2026flashforward} avoid dense clean updates only by training the model to condition on noisy context or to split planner and renderer roles, sacrificing drop-in compatibility with frozen base weights and LoRAs. On frozen checkpoints, ARCache~\cite{nan2026arcache} curbs cross-segment step-caching drift but leaves the persistent KV cache untouched. The natural training-free alternative, binary token reuse~\cite{dutson2023eventful,bolya2023tome}, refreshes only some tokens and keeps stale KV for the rest. Under early exit, however, the update and the denoiser form a \emph{closed loop}: stale KV corrupts the conditioning of subsequent blocks, delays their token convergence, and inflates downstream denoising beyond that of dense updates (Fig.~\ref{fig:teaser}c), while the accumulated error surfaces as the late-horizon ghosting of Fig.~\ref{fig:qualitative}. Unstructured token selection further incurs GPU wave-quantization and host-side routing overhead~\cite{dao2022flashattention,liu2023efficientvit,liang2026vmonarch}.

Our key observation is that most clean KV need not be recomputed, because two complementary estimates of it already reside in GPU memory before the update: (i)~the historical KV of the motion-matched patch in the previous block, which carries no current-step bias and is position-aligned once explicit 3D RoPE~\cite{su2024roformer} phases are removed, but becomes stale under disocclusion; and (ii)~the in-flight KV each token holds at its exit step, which reflects the current content up to a step-dependent modulation shift and residual noise. Since this shift is largely determined by the exit step, offline-calibrated per-head affine maps in Pre-RoPE space remove it, and the two estimates can be fused without extra DiT forwards. True $t=0$ refreshes are then needed only for tokens whose errors would change future attention readouts, such as attended disocclusions; an offline readout proxy identifies them on both LongLive~\cite{yang2025longlive} and Self-Forcing~\cite{huang2025selfforcing}.

Building on this observation, we introduce \textbf{\method} (\emph{Step-Aware Memory Reconstruction}):
\begin{itemize}[leftmargin=*,nosep]
\item \textbf{Step-aware Pre-RoPE memory reconstruction.} We reconstruct unrefreshed KV in phase-aligned Pre-RoPE space, pairing exit-step-conditioned affine maps with a clipped excess-error fusion rule that keeps static tokens on clean history while tracking newly exposed content.
\item \textbf{Readout-guided refresh selection with efficient execution.} We derive a first-order attention-readout error proxy with a staleness term, and co-design it with tile-aligned plans, an NNLS latency model, and synchronization-free Triton kernels ($0.33\%$ DiT overhead).
\item \textbf{Closed-loop and cross-backbone validation.} On 48-second LongLive-1.3B videos, \method reduces DiT latency by \FullLatencyReduction{}\% (\FullSpeedup{}$\times$) and principal FLOPs by \FullTFLOPCut{}\% vs.\ dense KV updates while improving Q4 PSNR over reuse by $+\PSNRGain{}$\,dB ($+1.05$\,dB at matched latency; $+\UntouchedGain{}$\,dB on held-out VidProM-16). On frozen Self-Forcing-DMD-1.3B, the reconstruction mitigates reuse-induced step inflation ($\SFDReuseSteps{} \to \SFDOursSteps{}$) and improves late PSNR ($+\SFDPSNRGain{}$\,dB) zero-shot, restores diagnostic quality (\SFDIndomainVB{}) after a one-time training-free calibration, and reaches \SFDCFDynPSNR{}\,dB / \SFDCFDynVB{} with chained KV updates and adaptive budgeting.
\end{itemize}

\section{Related Work}
\label{sec:related}

\begin{figure*}[!t]
\centering
\includegraphics[width=0.98\textwidth]{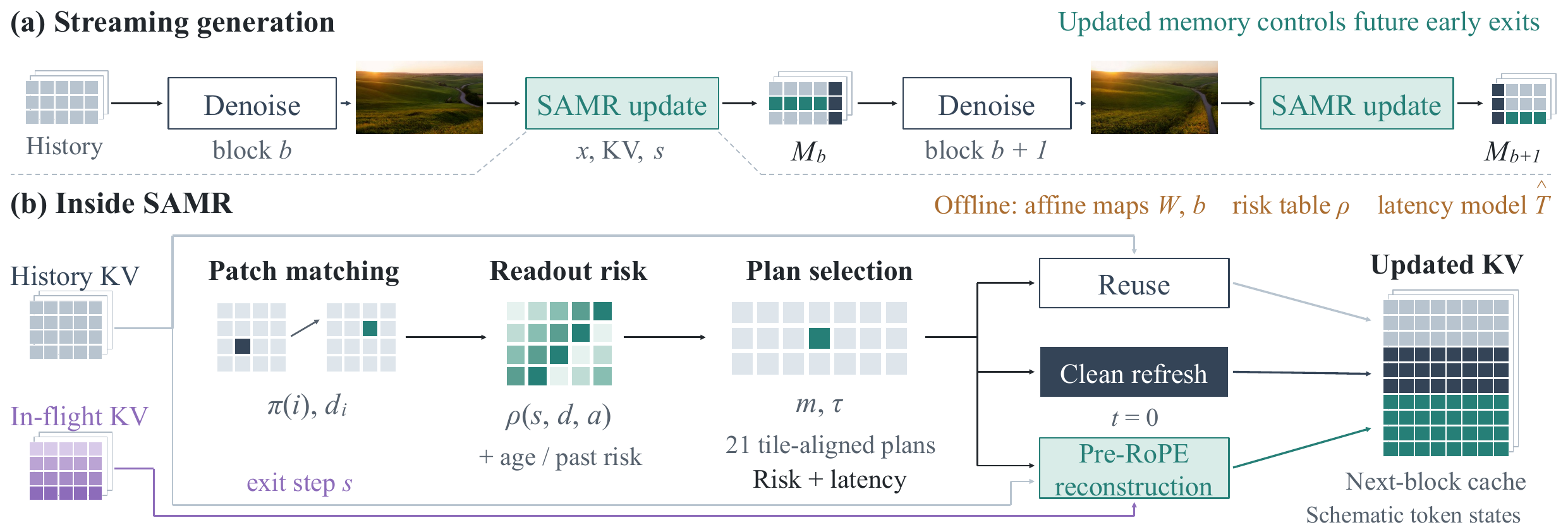}
\caption{\textbf{Overview of \method.} \textbf{(a)} Streaming loop: denoising block $b$ yields latents, in-flight KV, and exit steps $s$; \method then writes the cache $M_b$ read by block $b{+}1$, so memory fidelity sets future early exits and closes the loop of Eq.~\eqref{eq:closed-loop-main}. \textbf{(b)} One-time offline calibration fits the exit-step-conditioned Pre-RoPE affine maps $(W_{\ell s h}, b_{\ell s h})$, the readout-risk table $\rho$, and the KV-update + Step-1 latency model $\widehat{T}$; nothing is fitted at test time. Per block, online control runs register-resident local patch matching ($\pi(i)$, $d_i$), a readout-risk lookup by exit step, distance, and action $a$ with staleness, and single-wave scoring of the 21 tile-aligned plans $\mathcal{P}_b$ (kernels \CorrSpeedup{}$\times$ and \PlanSpeedup{}$\times$ faster than PyTorch). The KV update applies phase-aligned reuse on $\mathcal{U}_b$, $t=0$ refresh on $\mathcal{R}_b$, and Pre-RoPE reconstruction on $\mathcal{G}_b$; refresh rows share one tile-aligned pass with Step~1 of block $b{+}1$. Token colors are schematic; thumbnails are \method frames (LongLive).}
\label{fig:overview}
\end{figure*}

\paragraph{Streaming video diffusion and KV caching.}
Bidirectional video DiTs~\cite{wan2025} can be distilled into block-causal generators with a bounded KV cache~\cite{yin2025causvid}; Self Forcing~\cite{huang2025selfforcing,yin2024dmd2} trains on self-rollouts to reduce exposure bias, and LongLive~\cite{yang2025longlive,su2024roformer,xiao2024streamingllm} reaches minute-long videos with cache-relative RoPE and attention sinks. TriTD~\cite{li2026tritd} (noisy context) and FlashForward~\cite{wang2026flashforward} (planner--renderer roles) change cache production through training, giving up drop-in compatibility with frozen weights and LoRAs. Among training-free engines, Speculative Forcing~\cite{choi2026speculative} stops converged tokens early and merges the clean KV update with the next block's first step; Deep Forcing~\cite{yi2026deepforcing}, Forcing-KV~\cite{ji2026forcingkv}, FAST-AR~\cite{samuel2026fastar}, and Light Forcing~\cite{lv2026lightforcing} compress or sparsify the cache that is \emph{read}; and Commit Forcing~\cite{jing2026commit} restricts the clean update to a chained local window. \method is complementary to these engines: it reduces the remaining \emph{write-side} cost by refreshing fewer tokens and reconstructing the rest.

\paragraph{Diffusion caching and hardware-aware design.}
Diffusion caching reuses computation across tokens or timesteps, via token merging and reuse~\cite{bolya2023tome,dutson2023eventful} or step caches guided by input change (TeaCache~\cite{liu2025teacache}), sensitivity (SenCache~\cite{haghighi2026sencache}), spectral content (SeaCache~\cite{chung2026seacache}), attention broadcasting (PAB~\cite{zhao2025pab}), token-level selection (ToCa~\cite{zou2025toca}), feature forecasting (TaylorSeer~\cite{liu2025taylorseer}), or shallow probes (DiCache~\cite{bu2026dicache}). For streaming video, ARCache~\cite{nan2026arcache} schedules step caching from history and corrects the resulting drift. These methods skip computation \emph{within} a block, whereas \method maintains the persistent cache passed \emph{across} blocks, where tokens exit at heterogeneous timesteps. Following hardware-aware designs~\cite{dao2022flashattention,liu2023efficientvit,liang2026vmonarch,liu2025cachequant}, we align sparse work to GPU tiles and fuse it into few kernels so that the reductions translate into wall-clock speedups.

\section{Problem Formulation and Method}
\label{sec:formulation}
\label{sec:method}

\subsection{Preliminaries: Clean KV Updates and Closed-Loop Cost}
\label{sec:preliminaries}
\paragraph{Block-causal generation and clean KV updates.}
Block-causal video diffusion~\cite{yin2025causvid,huang2025selfforcing,yang2025longlive} generates sequential latent blocks $b \in \{0, \dots, B-1\}$ of $N$ spatiotemporal tokens, conditioned on a bounded persistent KV cache $M_{<b}$ across $L$ DiT layers ($H$ heads of dimension $d_h$). With frozen weights $\theta$, each block alternates two operators:
(1)~\textbf{Denoising} $x_b = \mathcal{D}_\theta(\epsilon_b, c, M_{<b})$ runs up to $S$ steps $t_{S-1} > \dots > t_0 > 0$. Under speculative early exit~\cite{choi2026speculative}, token $i$ halts at step $s_i \in \{0,\dots,S-1\}$ once converged, leaving noisy \emph{in-flight KV} $(K_{b,i}^{\mathrm{if}}, V_{b,i}^{\mathrm{if}})$ computed at $t_{s_i} > 0$.
(2)~\textbf{Clean KV update} $M_b = \mathcal{U}_\theta(x_b, M_{<b})$ runs a noise-free ($t=0$) forward pass on the finished block and writes clean targets $Y_{b,i} = (K_{b,i}^{\mathrm{clean}}, V_{b,i}^{\mathrm{clean}})$ into the cache for all future blocks $b' > b$. Caching in-flight KV directly would instead corrupt autoregressive conditioning through residual noise and heterogeneous $t_{s_i}$-dependent AdaLN shifts. Deployed engines~\cite{choi2026speculative,jing2026commit} restrict $\mathcal{U}_\theta$ to a chained local window and merge it with Step~1 of block $b{+}1$. Fig.~\ref{fig:overview}a depicts this loop, in which the cache written after block $b$ conditions the denoising of all later blocks; \method replaces the dense $t=0$ pass with the sparse update of Fig.~\ref{fig:overview}b.

\paragraph{Closed-loop coupling under sparse refresh.}
Suppose the $t=0$ pass refreshes only a subset $\mathcal{R}_b$ of $m_b = |\mathcal{R}_b| \le N$ tokens and fills the remaining $N - m_b$ entries without a DiT forward, giving an approximate cache $\widehat{M}_b$. The total streaming cost then couples the write-side update with read-side early-exit denoising:
\begin{equation}
 \mathcal{C}_{\mathrm{tot}} = \sum_{b=0}^{B-1} \Bigl[ \mathcal{C}_{\mathrm{den}}\bigl(x_b \,\big|\, \widehat{M}_{<b}\bigr) + \mathcal{C}_{\mathrm{upd}}\bigl(\widehat{M}_b \,\big|\, x_b, \widehat{M}_{<b}\bigr) \Bigr].
\label{eq:closed-loop-main}
\end{equation}
The terms are coupled because later exit steps $s_{b',j}(\widehat{M}_{<b'})$ depend on cache fidelity: copying stale KV reduces $\mathcal{C}_{\mathrm{upd}}$ but delays future convergence and increases $\mathcal{C}_{\mathrm{den}}$. Nor is $s_i$ a reliable refresh criterion: it reflects within-block convergence, not novelty relative to the cache (an entering rigid object may converge quickly yet be absent from $\widehat{M}_{<b}$), and on the eight calibration prompts (\emph{Dev-8}) within-block residuals correlate only moderately with between-block change ($\rho{=}0.643$). Frozen-weight acceleration thus requires (i)~a forward-free estimator of the clean KV for $i \notin \mathcal{R}_b$ (\S\ref{sec:transport}), (ii)~a criterion for choosing $\mathcal{R}_b$ (\S\ref{sec:risk}), and (iii)~an execution path whose overhead does not cancel the savings (\S\ref{sec:system}).

\begin{algorithm}[!t]
\footnotesize
\caption{\method KV update for block $b$}
\label{alg:samr}
\begin{algorithmic}[1]
\Require Finished block $x_b$, last frame $x_{b-1}^{\mathrm{last}}$, in-flight KV $\{K_{b,i}^{\mathrm{if}}, V_{b,i}^{\mathrm{if}}\}$ with exit steps $\{s_i\}$, cache $\widehat{M}_{<b}$, staleness state $(h_{\mathrm{prev}}, u_{\mathrm{prev}})$
\Ensure Updated cache $\widehat{M}_b$ and staleness state $(h, u)$
\State $(\pi_i, d_i)_{i=1}^N \leftarrow \textsc{LocalPatchMatch}(x_b, x_{b-1}^{\mathrm{last}})$
\State Bin $c_i \leftarrow \phi(s_i, d_i)$; compute risks $\widehat{r}_i(a)$ from $\rho_{c_i}(a)$ and priority $\psi_i$
\State Solve Eq.~\eqref{eq:objective} over $\mathcal{P}_b$ in one wave; split tokens into $\mathcal{R}_b \sqcup \mathcal{G}_b \sqcup \mathcal{U}_b$
\For{layer $\ell = 0, \dots, L-1$}
  \State Remove RoPE on $\mathcal{G}_b \cup \mathcal{U}_b$ via Eq.~\eqref{eq:unrope} $\to (X_i^{\mathrm{old}}, X_i^{\mathrm{cur}})$
  \State On $\mathcal{G}_b$: correct via Eq.~\eqref{eq:transport}, fuse via Eq.~\eqref{eq:gain}, restore norm
  \State Re-apply RoPE $R(p_{b,i})$ on $\mathcal{G}_b \cup \mathcal{U}_b$; write to $\widehat{M}_b$
  \State Merged $(|\mathcal{R}_b| + n_1)$-row pass: $t=0$ refresh on $\mathcal{R}_b$ + Step~1 of block $b{+}1$
\EndFor
\State Reset $h_i \leftarrow 0$ on $\mathcal{R}_b$; update $(h_i, u_i)$ on $\mathcal{G}_b \cup \mathcal{U}_b$; \Return $(\widehat{M}_b, h, u)$
\end{algorithmic}
\end{algorithm}

\begin{table*}[!t]
\centering
\small
\setlength{\tabcolsep}{2.7pt}
\begin{tabular}{@{}lrrrrrrrrrrrr@{}}
\toprule
 & \multicolumn{6}{c}{DiT Execution, Compute \& Memory Actions} & \multicolumn{4}{c}{Ref.-Trajectory Agreement (Q4)} & \multicolumn{2}{c}{No-Ref.\ CLIP} \\
\cmidrule(lr){2-7}\cmidrule(lr){8-11}\cmidrule(lr){12-13}
Method & DiT $\downarrow$ & Spd. $\uparrow$ & TFLOPs $\downarrow$ & Steps $\downarrow$ & Ref. & Rec. & PSNR $\uparrow$ & Sat. $\downarrow$ & Flk. $\downarrow$ & Mot. $\downarrow$ & Anc. $\uparrow$ & Text $\uparrow$ \\
 & (s) & vs.\ Full & /vid & /tok & (\%) & (\%) & (dB) & (\%) & (\%) & (\%) & (CLIP) & (CLIP) \\
\midrule
\multicolumn{13}{@{}l}{\textit{(a) Primary 48-second suite (\texttt{Main 15$\times$3}: 45 paired rollouts, LongLive-1.3B, U-shaped BF16 engine)}} \\
Full memory (reference) & 34.71 & 1.00$\times$ & 4041 & 1.78 & 100.0 & 0.0 & -- & -- & -- & -- & \textbf{0.877} & 0.334 \\
Eventful reuse & 30.13 & 1.15$\times$ & 3242 & 1.87 & 24.6 & 0.0 & 13.87 & 11.17 & 23.82 & 23.03 & 0.856 & 0.334 \\
Calibrated assimilation & 29.43 & 1.18$\times$ & 3013 & 1.72 & 24.6 & 38.9 & 14.92 & \textbf{7.99} & \textbf{20.63} & \textbf{18.73} & 0.860 & 0.337 \\
\rowcolor{gray!10}
\textbf{SAMR (ours)} & \textbf{27.47} & \textbf{1.26$\times$} & \textbf{2799} & \textbf{1.58} & 24.7 & 45.2 & \textbf{15.47} & 8.70 & 23.46 & 22.81 & 0.865 & \textbf{0.340} \\
\midrule
\multicolumn{13}{@{}l}{\textit{(b) External 48-second validation (\texttt{VidProM-16}: 16 held-out clips, fresh seed 880021, locked policy)}} \\
Full memory (reference) & 36.71 & 1.00$\times$ & 4337 & 1.97 & 100.0 & 0.0 & -- & -- & -- & -- & \textbf{0.870} & 0.346 \\
Eventful reuse & 32.74 & 1.12$\times$ & 3612 & 2.12 & 24.6 & 0.0 & 12.64 & 11.71 & 26.36 & 24.45 & 0.858 & 0.349 \\
Calibrated assimilation & 31.92 & 1.15$\times$ & 3357 & 1.95 & 24.6 & 45.1 & 13.45 & 7.87 & 19.54 & 18.62 & 0.853 & 0.345 \\
\rowcolor{gray!10}
\textbf{SAMR (ours, locked)} & \textbf{30.41} & \textbf{1.21$\times$} & \textbf{3152} & \textbf{1.82} & 23.6 & 54.0 & \textbf{13.78} & \textbf{6.68} & \textbf{17.99} & \textbf{17.89} & 0.862 & \textbf{0.352} \\
\bottomrule
\end{tabular}

\caption{\textbf{Closed-loop efficiency and quality on 48-second LongLive-1.3B videos} (64 blocks / 765 frames, $480\times832$, AMD GPUs (80 CUs each); all methods share the accelerated engine of \S\ref{sec:setup}). \emph{TFLOPs}: trace-exact principal DiT compute. \emph{Steps}: mean denoising steps per token. \emph{Ref.}/\emph{Rec.}: shares of refreshed / reconstructed tokens. \emph{Q4} columns measure agreement of the final quarter (36--48\,s) with the same-seed \emph{Full memory} output (\emph{Sat.}, \emph{Flk.}, \emph{Mot.}: saturation, flicker, and motion drift); \emph{Anc.}/\emph{Text} are CLIP similarities to early frames and the prompt.}
\label{tab:main}
\end{table*}

\subsection{Step-Aware Pre-RoPE Reconstruction}
\label{sec:transport}
\paragraph{Two complementary estimates.}
Before $\mathcal{U}_\theta$ runs on block $b \ge 1$, two approximations of $Y_{b,i}$ are already available at no forward cost:
(i)~\emph{historical KV} $(K_{b-1,\pi(i)}, V_{b-1,\pi(i)})$ of the matched patch $\pi(i) = \argmin_{j \in \mathcal{N}(i)} \norm{x_{b,i} - x_{b-1,j}^{\mathrm{last}}}_2$ in a local neighborhood of the previous block's last frame (normalized match distance $d_i$), which is free of $t_{s_i}>0$ bias but becomes stale under disocclusion; and
(ii)~\emph{in-flight KV} $(K_{b,i}^{\mathrm{if}}, V_{b,i}^{\mathrm{if}})$ at exit step $s_i$, which follows the geometry of $x_b$ up to an $s_i$-dependent modulation shift and residual noise.
\emph{Dev-8} probes support this design: $5\times5$ matching cuts MSE by $18.5\%$ over same-position copy, and history is more accurate in shallow layers and static patches, whereas corrected in-flight KV overtakes it for $\ell \ge 8$, $d_i \ge 0.35$, $s_i \ge 2$ on both backbones.

\paragraph{Pre-RoPE alignment.}
Cached keys already carry 3D rotary phases $R(p)$~\cite{su2024roformer,yang2025longlive}, so blending keys directly would conflate content differences with the positional offset $p_{b,i} - p_{b-1,\pi(i)}$. We therefore remove the phases with the conjugate rotation $R(p)^{*} = R(p)^{-1}$:
\begin{equation}
\begin{split}
 X_i^{\mathrm{old},K} &= R(p_{b-1,\pi(i)})^{*} K_{b-1,\pi(i)}, \\
 X_i^{\mathrm{cur},K} &= R(p_{b,i})^{*} K_{b,i}^{\mathrm{if}}.
\end{split}
\label{eq:unrope}
\end{equation}
Values are phase-free ($X_i^{\cdot,V} = V_\cdot$); writes re-apply $R(p_{b,i})$.

\paragraph{Exit-step-conditioned affine correction.}
With frozen $\theta$, the AdaLN-Zero modulation and noise variance of in-flight KV shift systematically with the exit step, so we remove this bias with offline per-layer, per-step, per-head affine maps:
\begin{equation}
 \widetilde{X}_i^{\mathrm{cur}} = X_i^{\mathrm{cur}} W_{\ell s_i h} + b_{\ell s_i h},
\label{eq:transport}
\end{equation}
which are fitted in closed form by identity-centered ridge regression on \emph{Dev-8} (Appendix Eq.~\ref{eq:ridge}); inference requires only one $d_h \times d_h$ projection per head.

\paragraph{Clipped excess-error fusion.}
We bin tokens by $c_i = \phi(s_i, d_i)$ and estimate offline, per bin, the Pre-RoPE error moments $e_o = \E[\norm{X^{\mathrm{old}} - Y}_2^2]$, $e_r = \E[\norm{\widetilde{X}^{\mathrm{cur}} - Y}_2^2]$, and $c_{or} = \E[\langle X^{\mathrm{old}} - Y, \widetilde{X}^{\mathrm{cur}} - Y\rangle]$. Minimum-variance rules, whether diagonal, $e_o/(e_o + e_r)$, or correlated-residual BLUE, $(e_o - c_{or})/(e_o + e_r - 2c_{or})$, assign nonzero weight to the noisy in-flight estimate even on static patches where history is already more accurate, injecting jitter that perturbs future early exit (\S\ref{sec:ablation-text}). We instead clip the excess error:
\begin{align}
 w_{\ell c h} &= \frac{(e_o - e_r)_+}{(e_o - e_r)_+ + e_r + \varepsilon}, \nonumber\\
 Z_i &= (1 - w_{\ell c_i h}) X_i^{\mathrm{old}} + w_{\ell c_i h} \widetilde{X}_i^{\mathrm{cur}}.
\label{eq:gain}
\end{align}
The hinge vanishes whenever $e_o \le e_r$, locking static tokens to clean history, and the weight rises smoothly toward $\widetilde{X}_i^{\mathrm{cur}}$ under disocclusion ($e_o > e_r$). Tokens with $d_i < \tau_d$ and $s_i < \tau_s$ skip Eq.~\eqref{eq:transport} and reuse history at zero FLOPs; the rest are norm-restored (Appendix Eq.~\ref{eq:spherical}) before write-back.

\subsection{Readout-Guided Refresh Selection}
\label{sec:risk}
\label{sec:decision}
\paragraph{First-order readout proxy.}
KV error matters only insofar as it changes what future queries read. For query $q \in \R^{d_h}$, attention output $o = \sum_j p_j v_j$ and $p_j = \operatorname{softmax}_j(q^\top k_j / \sqrt{d_h})$, a first-order expansion with softmax coupling gives the effect of a single-token error $(\delta k_i, \delta v_i)$:
\begin{equation}
 \delta o_i
 \approx p_i \left[ \delta v_i + \frac{q^\top \delta k_i}{\sqrt{d_h}} (v_i - o) \right].
\label{eq:readout}
\end{equation}
It suppresses unattended tokens ($p_i \approx 0$) and emphasizes attended disocclusions ($v_i - o$). As block-$(b{+}1)$ queries are unavailable when writing block $b$, we tabulate the Empirical-Bayes-smoothed expected readout energy $\rho_c(a)$ offline per bin $c$ and action $a \in \{\mathrm{reuse}, \mathrm{reconstruct}\}$ (Appendix Eqs.~\ref{eq:risk}--\ref{eq:normalized-risk}). Online, each token inherits age $h_i = h_{\mathrm{prev}}[\pi_i] + 1$ and accumulated risk $u_i = u_{\mathrm{prev}}[\pi_i]$ from its match, giving the staleness-penalized risk $\widehat{r}_i(a) = \rho_{c_i}(a)(1 + \gamma_h \min(h_i, H_{\max})) + \gamma_u^{(a)} u_i$ (Appendix Eq.~\ref{eq:lineage}), so repeatedly reconstructed entries are refreshed sooner. A refresh resets $h_i \leftarrow 0$ and leaves residual risk $\eta \min_a \widehat{r}_i(a)$.

\paragraph{Tile-aligned plan selection.}
Tokens are ordered once by priority $\psi_i$ (Appendix~A), and \method evaluates a small grid $\mathcal{P}_b = \mathcal{M} \times \mathcal{C}$ of plans $\Pi_b = (m, \tau)$, where refresh counts $m \in \mathcal{M}$ are multiples of the GPU tile size $\Delta_m$ and $\tau \in \mathcal{C}$ are gain thresholds. A plan refreshes the top-$m$ tokens ($\mathcal{R}_b$), reconstructs tokens with gain $\widehat{r}_i(\mathrm{reuse}) - \widehat{r}_i(\mathrm{reconstruct}) > \tau \bar{g}_b$ ($\mathcal{G}_b$), and reuses the rest ($\mathcal{U}_b$; Alg.~\ref{alg:samr}). We select
\begin{equation}
 \Pi_b^* = \argmin_{\Pi_b \in \mathcal{P}_b}
 \left\{ \frac{1}{N}\sum_{i=1}^N \widehat{r}_i\bigl(a_i(\Pi_b)\bigr)
 + \lambda \frac{\widehat{T}(\Pi_b)}{T_0} \right\},
\label{eq:objective}
\end{equation}
where $\widehat{T}$ is an offline NNLS latency model of the merged KV-update + Step-1 pass. Because refreshed queries attend over the cache already written for $\mathcal{G}_b \cup \mathcal{U}_b$, refresh boundaries co-adapt to their reconstructed neighbors ($27.49$ vs.\ $27.28$\,dB at $m{=}0.25N$ on \emph{Dev-8}).

\section{Hardware-Aligned Execution}
\label{sec:system}

\begin{figure*}[!t]
\centering
\includegraphics[width=0.98\textwidth]{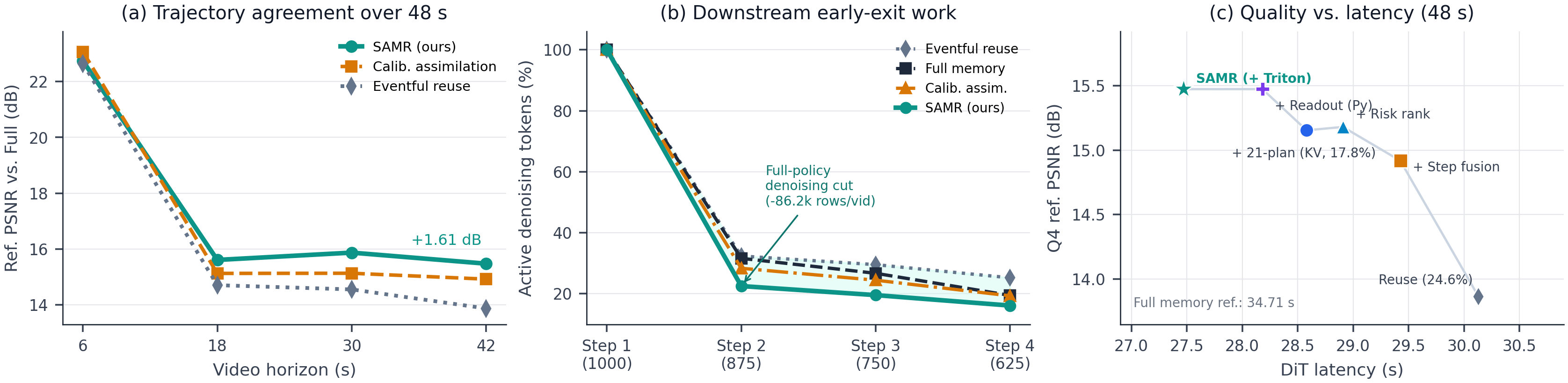}
\caption{\textbf{Closed-loop effect of each component on 48-second videos (\texttt{Main 15$\times$3}).} \textbf{(a)} PSNR against the same-seed \emph{Full memory} output per 12-second quarter. \textbf{(b)} Active tokens at Steps~1--4. Reuse keeps \emph{more} tokens active than \emph{Full memory}, whereas \method keeps fewer (86.2k fewer denoising rows per video than reuse). \textbf{(c)} Q4 PSNR versus DiT latency as components are added; the two plan-selection variants refresh $17.8\%$ and $24.7\%$ of tokens, respectively.}
\label{fig:fidelity}
\end{figure*}

The decisions of \S\ref{sec:method} pay off only if executing them is cheaper than the update they replace; we remove three overheads.

\paragraph{Host-side routing.}
Eager Python routing launches many micro-kernels and host--device synchronizations per block~\cite{liu2023efficientvit}. We replace it with two synchronization-free Triton kernels that reproduce the reference decisions exactly. \texttt{LocalPatchMatch} holds each query patch in registers and sweeps all $|\mathcal{N}(i)|$ candidates in an unrolled loop without HBM gathers. The plan-scoring kernel evaluates every plan in $\mathcal{P}_b$ within \emph{one GPU wave}, streaming risk tiles through SRAM and accumulating per-plan objectives (Eq.~\ref{eq:objective}) and step counts in registers, without materializing $|\mathcal{P}_b| \times N$ masks.

\paragraph{Cache memory and copies.}
On \emph{Dev-8}, the correlation between patch change and $t=0$ KV change is U-shaped over depth ($0.906 / 0.346 / 0.895$ at $\ell{=}0 / 15 / 29$). We exploit this by keeping a longer zero-copy ring buffer of $W_{\mathrm{bnd}}$ frames (sink plus rolling frames) in boundary layers $\mathcal{L}_{\mathrm{bnd}}$ and a shorter window $W_{\mathrm{mid}} < W_{\mathrm{bnd}}$ in middle layers $\mathcal{L}_{\mathrm{mid}}$. This \emph{U-shaped cache} cuts KV memory by $25.0\%$ and avoids per-step \texttt{torch.cat} copies~\cite{dao2022flashattention}. During the KV update (Alg.~\ref{alg:samr}), one fused Triton kernel per layer performs RoPE removal, affine correction, excess-error fusion, norm restoration, RoPE re-application, and BF16 write-back on $\mathcal{G}_b \cup \mathcal{U}_b$.

\paragraph{GEMM wave quantization.}
The $m^*$ refresh rows in $\mathcal{R}_b$ are concatenated with the $n_1$ Step-1 tokens of block $b{+}1$ so that both share the QKVO and FFN GEMMs~\cite{choi2026speculative,jing2026commit}. Arbitrary row counts $(m^* + n_1)$, however, trigger wave-occupancy fallbacks on multi-CU GPUs. Snapping both $\mathcal{M}$ and the active early-exit counts to multiples of $\Delta_m$ preserves dense Matrix-Core utilization (\S\ref{sec:absolute}). Optional low-bit quantization and causal VAE pipelining are in Appendix~D.

\section{Experiments}
\label{sec:experiments}

We ask whether \method shortens the KV update \emph{without} inflating downstream denoising (\S\ref{sec:absolute}), whether this generalizes (\S\ref{sec:sfd}), and which components matter (\S\ref{sec:ablation-text}).

\subsection{Experimental Setup}
\label{sec:setup}
\paragraph{Models and engines.}
We evaluate two frozen causal video DiTs built on Wan2.1-T2V-1.3B~\cite{wan2025} ($L{=}30$ layers, $D{=}1536$, $H{=}12$ heads, $d_h{=}128$, $N{=}4{,}680$ tokens per 3-frame block at $480\times832$, $S{=}4$ distilled steps): \textbf{LongLive-1.3B}~\cite{yang2025longlive} on 48-second streams ($64$ blocks / $765$ frames at $16$\,FPS) and \textbf{Self-Forcing-DMD-1.3B}~\cite{huang2025selfforcing} on 12-second streams ($16$ blocks / $189$ frames). The main 48-second runs (Tabs.~\ref{tab:main} and~\ref{tab:ablation}) execute on AMD GPUs, each with 80 compute units (CUs), using a BF16 engine that already stacks 4-step distillation~\cite{yin2024dmd2}, speculative early exit~\cite{choi2026speculative}, the U-shaped cache of \S\ref{sec:system} ($W_{\mathrm{bnd}}{=}12$ frames on $\ell \in \{0..5,24..29\}$, $W_{\mathrm{mid}}{=}7$ on $\ell \in \{6..23\}$; KV memory $3{,}290.6 \to 2{,}468.0$\,MiB), chained two-block KV updates~\cite{jing2026commit}, and merged KV-update + Step-1 GEMMs. All baselines share this engine. Tabs.~\ref{tab:quality-transfer} and~\ref{tab:est-ablation} use an earlier \emph{uniform-cache engine} (12-frame BF16 cache; Appendix~E).

\paragraph{Compared methods.}
\emph{Full memory} (dense) refreshes all tokens at $t{=}0$ (quality reference). \emph{Eventful reuse}~\cite{dutson2023eventful} refreshes a fixed $24.6\%$ of tokens, ranked by cumulative patch change, keeping stale KV elsewhere. \emph{Calibrated assimilation} is our reconstruction at this budget without plan selection; \method adds readout-guided plans and Triton kernels.

\paragraph{Calibration and hyperparameters.}
Offline calibration uses only the eight \emph{Dev-8} prompts and yields $90.7$\,MiB of BF16 Pre-RoPE tables ($47.55$\,M scalars) with $5\times5$ patch matching and $20$ $(s_i,d_i)$ bins. We use skip thresholds $(\tau_d{=}0.22, \tau_s{=}2)$, residual risk $\eta{=}0.1$, a $\Delta_m{=}256$-aligned grid of $21$ plans ($(m,\tau) \in \{512,\dots,2304\} \times \{0, 0.5, 1.0\}$), and a 9-feature NNLS latency model ($T_0{=}100$\,ms, $\lambda{=}1.0$; test MAE \CostModelMAE{}\,ms / \CostModelRelErr{}\%, $r{=}0.943$). One denoising token costs $5.18$\,GFLOPs and one chained $t=0$ update token $4.32$\,GFLOPs ($20.2$\,TFLOPs/block; $+258.5$\,ms/block marginal), whereas the per-head Pre-RoPE projection costs $23.59$\,MFLOPs/token ($183\times$ fewer principal FLOPs, an arithmetic ratio rather than a wall-clock speedup).

\paragraph{Benchmarks and metrics.}
\texttt{Main 15$\times$3} ($15$ prompts $\times$ $3$ seeds, $45$ noise-paired rollouts) is the primary benchmark. \texttt{VidProM-16} ($16$ VidProM~\cite{wang2024vidprom} prompts, 48\,s, fresh seed $880021$) validates the locked policy; its prompts come from a 64-prompt 12-s pool excluded from all fitting and screening. We report two metric families (Appendix~C):
(1)~\emph{Reference-trajectory agreement} with the same-seed \emph{Full memory} output: quarterly PSNR, SSIM~\cite{wang2004ssim}, normalized AlexNet distance (\emph{AlexD.}: LPIPS-style~\cite{zhang2018lpips} \texttt{relu1}--\texttt{relu5} distance without learned weights; lower is better), and saturation, flicker, and motion drift. We focus on the final quarter (\emph{Q4}, 36--48\,s).
(2)~\emph{Reference-free diagnostics}: CLIP ViT-B/32~\cite{radford2021clip,schuhmann2022laion5b} similarity to the prompt and to early frames, Farneback~\cite{farneback2003flow} smoothness and static stability, subject consistency, and a VBench-inspired~\cite{huang2024vbench} composite \emph{Diag.} ($0.25$ subject $+\,0.25$ smoothness $+\,0.20$ flicker $+\,0.15$ text $+\,0.15$ anti-drift; Appendix Eq.~\eqref{eq:diagcomp}). Neither replaces standard LPIPS or VBench-Long.
Long-horizon PSNR is trajectory-sensitive: perturbing the block-1 noise by $\epsilon \in \{10^{-3}, 10^{-2}\}$ under $100\%$ dense KV updates already lowers 16-block $\text{PSNR}_{\text{all}}$ to \FloorEpsSmallPSNR{}--\FloorEpsLargePSNR{}\,dB at unchanged \emph{Diag.} ($93.84$--$94.01$; Tab.~\ref{tab:est-ablation}, top). We thus read PSNR as trajectory fidelity, jointly with drift and visual inspection.

\begin{table*}[!t]
\centering
\small
\setlength{\tabcolsep}{3.0pt}
\begin{tabular}{@{}lrrrrrrrrr@{}}
\toprule
 & \multicolumn{3}{c}{Execution} & \multicolumn{3}{c}{No-Ref.\ Quality Diagnostics} & \multicolumn{3}{c}{Ref.-Trajectory Agreement} \\
\cmidrule(lr){2-4}\cmidrule(lr){5-7}\cmidrule(lr){8-10}
Method & DiT $\downarrow$ & Steps $\downarrow$ & Ref. & Diag. $\uparrow$ & Subj. $\uparrow$ & Smooth. $\uparrow$ & SSIM $\uparrow$ & AlexD. $\downarrow$ & PSNR $\uparrow$ \\
 & (s) & /tok & (\%) & (comp.) & (\%) & (\%) & & (norm.) & (dB) \\
\midrule
\multicolumn{10}{@{}l}{\textit{(a) LongLive-1.3B, 48\,s / 765 frames (8 prompts, uniform-cache engine)}} \\
Full memory (12f ref.) & 43.00 & 1.73 & 100.0 & \textbf{93.56} & 90.25 & \textbf{97.40} & 1.000 & 0.000 & -- \\
Periodic dense reset ($K{=}4$) & 35.05 & 1.89 & 23.8 & 91.76 & 89.18 & 96.49 & 0.541 & 0.809 & 14.85 \\
Eventful reuse & 37.70 & 1.94 & 24.6 & 92.10 & 89.10 & 96.70 & 0.529 & 0.831 & 13.89 \\
Eventful reuse ($2\times$ budget) & 39.35 & 1.88 & 49.2 & 92.45 & 89.59 & 97.25 & 0.567 & 0.736 & 15.56 \\
\textbf{Pre-RoPE recon.\ (matched)} & 36.37 & 1.76 & 24.6 & 92.77 & 89.85 & \textbf{97.40} & 0.552 & 0.780 & 14.92 \\
+ Chained KV + Pre-RoPE recon. & \textbf{32.85} & \textbf{1.66} & 24.6 & 93.41 & \textbf{90.68} & 97.39 & \textbf{0.597} & \textbf{0.688} & \textbf{15.77} \\
\midrule
\multicolumn{10}{@{}l}{\textit{(b) Self-Forcing-DMD-1.3B, 12\,s / 189 frames (16 prompts, uniform-cache engine; cross-backbone transfer)}} \\
Full memory (reference) & 11.73 & 2.22 & 100.0 & \textbf{92.58} & \textbf{91.06} & 96.89 & 1.000 & 0.000 & -- \\
No clean KV update ($0\%$) & \textbf{8.59} & \textbf{2.13} & 0.0 & 90.48 & 90.11 & 95.59 & 0.548 & 0.881 & 14.45 \\
Eventful reuse & 10.65 & 2.47 & 24.6 & 91.16 & 89.87 & 96.99 & 0.577 & 0.817 & 14.28 \\
Eventful reuse ($2\times$ budget) & 10.96 & 2.38 & 49.2 & 91.71 & 90.40 & 97.01 & 0.616 & 0.691 & 15.25 \\
Pre-RoPE recon.\ (zero-shot \llcal{}) & 10.43 & 2.30 & 24.6 & 89.62 & 89.14 & 97.00 & 0.582 & 0.820 & 14.61 \\
Pre-RoPE recon.\ (in-domain calib.) & 10.43 & 2.31 & 24.6 & 91.10 & 90.14 & 97.15 & 0.587 & 0.798 & 14.77 \\
+ Chained KV + adaptive (stacked) & 9.69 & 2.15 & 30.5 & 92.10 & \textbf{91.16} & \textbf{97.18} & \textbf{0.631} & \textbf{0.654} & \textbf{16.32} \\
\bottomrule
\end{tabular}

\caption{\textbf{Reference-free diagnostics and trajectory agreement on the uniform-cache engine.} \emph{Subj.}/\emph{Smooth.}: subject consistency / smoothness. \textbf{(a)}~LongLive-1.3B, 48\,s, 8 prompts. \textbf{(b)}~Self-Forcing-DMD-1.3B, 12\,s, 16 prompts: zero-shot LongLive maps (\llcal{}), one-time training-free in-domain calibration, and stacking with chained KV updates + adaptive budgeting. Full tables: Appendix~E.}
\label{tab:quality-transfer}
\end{table*}

\subsection{Closed-Loop Efficiency and 48-Second Quality}
\label{sec:absolute}

\paragraph{Reuse moves the cost downstream.}
Tabs.~\ref{tab:main}a and~\ref{tab:ablation} and Fig.~\ref{fig:fidelity} trace each stage. Under \emph{Full memory} (\FullLatency{}\,s), the merged KV-update + Step-1 stage takes \FullCommitStepOneS{}\,s ($59.3\%$) and Steps~2--4 take \FullDenoiseStageS{}\,s (\FullSteps{} steps/token). \emph{Eventful reuse} skips $75.4\%$ of clean refreshes, cutting this stage to \ReuseCommitStepOneS{}\,s, but stale KV keeps $25$--$32\%$ of tokens active at Steps~2--4 (Fig.~\ref{fig:fidelity}b), inflating denoising to \ReuseSteps{} steps/token and \ReuseDenoiseStageS{}\,s. DiT latency falls only to \ReuseLatency{}\,s, at \ReusePSNR{}\,dB Q4 PSNR (ghosting in Fig.~\ref{fig:qualitative}).

\paragraph{Reconstruction reverses the inflation.}
At the same $24.6\%$ budget, \emph{Calibrated assimilation} reduces steps ($\ReuseSteps{} \to \LegacySteps{}$) and DiT time ($\ReuseLatency{} \to \LegacyLatency{}$\,s), raises Q4 PSNR to \LegacyPSNR{}\,dB, and attains the lowest mean Q4 saturation, flicker, and motion drift on \texttt{Main 15$\times$3} ($\LegacySat{}\%, \LegacyFlk{}\%, \LegacyMot{}\%$). Against a reuse baseline matched in latency on \emph{Dev-8} ($5\%$ tolerance: $30.15$ vs.\ $29.46$\,s), it retains $+1.05$\,dB Q4 PSNR.

\paragraph{Readout-guided plans and Triton execution.}
Selecting among 21 readout-scored plans per block costs \RoutingOld{}\,ms of eager routing per video; single-wave Triton control cuts it to \RoutingNew{}\,ms (\RoutingSpeedup{}$\times$; $0.33\%$ of DiT time) with exact decision parity ($5\times5$ matching \CorrSpeedup{}$\times$, plan scoring \PlanSpeedup{}$\times$). $\Delta_m{=}256$ alignment avoids up to $4.5\times$ off-tile GEMM slowdowns and reduces merged execution at $m{=}2{,}304$ from $654.1$ to $473.3$\,ms. The full \method reaches \MainSteps{} steps/token, \MainLatency{}\,s, and \MainPSNR{}\,dB Q4 PSNR, at the cost of higher mean Q4 flicker and motion drift than \emph{Calibrated assimilation} ($\MainFlk{}\% / \MainMot{}\%$); the latter therefore suits stability-critical use.

\begin{table}[!t]
\centering
\small
\setlength{\tabcolsep}{1.8pt}
\begin{tabular*}{\linewidth}{@{\extracolsep{\fill}}lrrrrr@{}}
\toprule
Variant & Steps & PSNR$_{\text{all}}$ & $\Delta$ & AlexD. & Diag. \\
 & /tok $\downarrow$ & (dB) $\uparrow$ & (dB) & $\downarrow$ & $\uparrow$ \\
\midrule
Dense $\epsilon{=}10^{-3}$ & 1.77 & 22.602 & -- & 0.289 & 94.01 \\
Dense $\epsilon{=}10^{-2}$ & 1.78 & 21.911 & -- & 0.297 & 93.84 \\
\midrule
Reuse (baseline) & 1.85 & 16.866 & -- & 0.608 & 93.29 \\
w/o age + reuse sel. & 1.96 & 16.575 & -0.463 & 0.631 & 92.48 \\
\midrule
Reference (with context) & 1.91 & 17.038 & 0.000 & 0.609 & 93.12 \\
\midrule
History only ($w \equiv 0$) & 2.22 & 16.077 & -0.961 & 0.671 & 92.04 \\
In-flight only ($w \equiv 1$) & 1.96 & 16.386 & -0.652 & 0.631 & \textbf{93.70} \\
w/o Pre-RoPE affine & 2.02 & 16.205 & -0.833 & 0.639 & 93.11 \\
Correlated BLUE gain & 2.02 & 16.295 & -0.743 & 0.660 & 92.72 \\
Unstratified $W, b$ & 1.93 & 16.762 & -0.276 & 0.629 & 93.08 \\
w/o spherical norm & 1.95 & 17.145 & +0.107 & 0.599 & 93.32 \\
\midrule
\rowcolor{gray!10}
\textbf{Simplified (w/o ctx)} & \textbf{1.79} & \textbf{17.325} & \textbf{+0.287} & \textbf{0.580} & 93.45 \\
\bottomrule
\end{tabular*}

\caption{\textbf{Estimator ablations and trajectory sensitivity on HeldOut-16} (LongLive-1.3B, 189 frames, uniform-cache engine). Dense perturbation controls use $100\%$ refresh; all other rows use $24.6\%$. History only uses matched selection. $\Delta$: $\text{PSNR}_{\text{all}}$ difference to the reference estimator. AlexD. is normalized; Diag. is composite. Full table: Appendix~E.}
\label{tab:est-ablation}
\vspace{-0.3em}
\end{table}

\begin{table}[!t]
\centering
\small
\setlength{\tabcolsep}{1.35pt}
\begin{tabular}{@{}lrrrrrr@{}}
\toprule
Variant & DiT$\downarrow$ & Route$\downarrow$ & Steps$\downarrow$ & Ref. & PSNR$\uparrow$ & Flk.$\downarrow$ \\
 & (s) & (ms) & /tok & (\%) & (dB) & (\%) \\
\midrule
Eventful reuse & 30.12 & 209 & 1.87 & 24.6 & 13.87 & 23.8 \\
+ Step fusion & 29.44 & 215 & 1.72 & 24.6 & 14.92 & 20.6 \\
+ Risk rank & 28.91 & 273 & 1.66 & 24.6 & 15.18 & \textbf{19.4} \\
+ 21-plan (KV) & 28.58 & 754 & 1.65 & 17.8 & 15.15 & 19.9 \\
+ Readout (Py) & 28.18 & 765 & 1.58 & 24.7 & 15.47 & 23.5 \\
\textbf{+ Triton (SAMR)} & \textbf{27.47} & \textbf{90} & \textbf{1.58} & 24.7 & \textbf{15.47} & 23.5 \\
\midrule
Cost $\lambda{=}2$ (Py) & 27.47 & 763 & 1.60 & 10.9 & 14.91 & 22.2 \\
\bottomrule
\end{tabular}

\caption{\textbf{Cumulative 48-second progression on \texttt{Main 15$\times$3}} (45 paired rollouts). \emph{Route}: per-video routing time. \emph{Step fusion}: fixed-budget \emph{Calibrated assimilation}; \emph{Risk rank}: risk-ranked refresh; \emph{21-plan}: per-block plan selection with a KV-error proxy (KV) or the readout risk (Readout), which select different refresh rates ($17.8\%$ vs.\ $24.7\%$).}
\label{tab:ablation}
\vspace{-0.3em}
\end{table}

\paragraph{Compute attribution and per-prompt consistency.}
Against \emph{Full memory}, \method reduces principal FLOPs by \FullTFLOPCut{}\% ($\FullTFLOPs{} \to \MainTFLOPs{}$\,TFLOPs) and DiT latency by \FullLatencyReduction{}\% (\FullSpeedup{}$\times$). Against reuse, it saves \DenoiseTFLOPSaved{}\,TFLOPs of downstream denoising ($\ReuseDenoiseStageS{} \to \MainDenoiseStageS{}$\,s), while the Pre-RoPE projection costs \MainReconTFLOPs{}\,TFLOPs ($0.11\%$ of FLOPs; saved-to-spent ratio $\approx \ReconFLOPRatio{}$). The gains are consistent: \method is faster than reuse on $14$ of $15$ prompts, by $2.66$\,s on average and by $9.18$\,s on handheld motion, where reuse is even slower than dense updates ($39.55$ vs.\ $38.78$\,s), and it attains higher Q4 PSNR on all $15$ ($+\PSNRGain{}$\,dB on average; Fig.~\ref{fig:qualitative}). Q4 saturation, flicker, and motion drift stay on par with reuse: the mean differences ($\MainSatDelta{}\%$, $\MainFlkDelta{}\%$, and $\MainMotDelta{}\%$) lie within prompt-to-prompt variation (Appendix Tab.~\ref{tab:paired}).

\paragraph{Reference-free quality.}
On the uniform-cache engine (Tab.~\ref{tab:quality-transfer}a), equal-budget ($24.6\%$) Pre-RoPE reconstruction improves over reuse in SSIM ($\AbsReuseSSIM{} \to \AbsOursSSIM{}$), \emph{AlexD.} ($\AbsReuseLPIPS{} \to \AbsOursLPIPS{}$), smoothness ($\AbsReuseSmooth{} \to \AbsOursSmooth{}$), and \emph{Diag.} ($\AbsReuseVB{} \to \AbsOursVB{}$). It also beats $2\times$-budget reuse in smoothness and \emph{Diag.} (dense: \AbsFullVB{}) while being $2.98$\,s faster, and with chained KV updates reaches \AbsCFDit{}\,s and \AbsCFVB{} \emph{Diag.}

\subsection{Generalization}
\label{sec:sfd}
\paragraph{Held-out 48-second prompts.}
On \texttt{VidProM-16} with the locked policy (Tab.~\ref{tab:main}b; Appendix Fig.~\ref{fig:untouched}), reuse inflates denoising ($\UntouchedFullSteps{} \to \UntouchedReuseSteps{}$ steps/token), whereas \method lowers it to \UntouchedSteps{} ($15/16$ prompts). It reduces TFLOPs by \UntouchedTFLOPCut{}\% and DiT latency from \UntouchedFullDit{} to \UntouchedDit{}\,s (\UntouchedSpeedup{}$\times$ vs.\ dense; faster than reuse on \UntouchedSpeedWins{}/16), and improves Q4 PSNR over reuse by $+\UntouchedGain{}$\,dB (\UntouchedPSNRWins{}/16 prompts) with lower mean saturation, flicker, and motion drift.

\paragraph{Second backbone: Self-Forcing-DMD-1.3B.}
On frozen Self-Forcing-DMD-1.3B~\cite{huang2025selfforcing} with 16 prompts (Tab.~\ref{tab:quality-transfer}b), omitting clean updates ($0\%$ refresh) degrades \emph{Diag.} ($\SFDFullVB{} \to 90.48$) and \emph{AlexD.} (\SFDNoCommitLPIPS{}), and $24.6\%$ \emph{Eventful reuse} again inflates denoising from \SFDFullSteps{} to \SFDReuseSteps{} steps/token (\SFDReuseDit{}\,s; \SFDReusePSNR{}\,dB final-quarter, or \emph{late}, PSNR). Against this baseline:
(1)~\emph{Zero-shot transfer (\llcal{}).} Unchanged LongLive tables reduce steps ($\SFDReuseSteps{} \to \SFDOursSteps{}$) and DiT time (\SFDOursDit{}\,s), and improve late PSNR by $+\SFDPSNRGain{}$\,dB ($\SFDReusePSNR{} \to \SFDOursPSNR{}$\,dB, $\SFDPSNRWins{}/16$ prompts). Modulation mismatch, however, lowers \emph{Diag.} ($\SFDReuseVB{} \to \SFDZeroshotVB{}$) and slightly raises \emph{AlexD.} ($\SFDReuseLPIPS{} \to \SFDOursLPIPS{}$).
(2)~\emph{One-time training-free calibration.} Refitting the closed-form ridge moments on Self-Forcing rollouts preserves the speed and step gains (\SFDOursDit{}\,s, $2.31$ steps/token) and recovers \emph{Diag.} to \SFDIndomainVB{}, surpassing reuse in subject consistency ($90.14\%$ vs.\ $89.87\%$), smoothness ($97.15\%$), \emph{AlexD.} (\SFDIndomainLPIPS{}), and late PSNR (\SFDIndomainPSNR{}\,dB), while remaining below dense updates (\SFDFullVB{}).
(3)~\emph{Stacked execution.} Adding chained KV updates and adaptive budgeting reduces DiT time to \SFDCFDynDit{}\,s ($1.21\times$ vs.\ dense \SFDFullDit{}\,s; $2.15$ steps/token, below dense \SFDFullSteps{}) and reaches \SFDCFDynPSNR{}\,dB PSNR, \SFDCFDynLPIPS{} \emph{AlexD.}, and \SFDCFDynVB{} \emph{Diag.}, beating $2\times$-budget reuse ($10.96$\,s) in speed and quality.

\subsection{Ablations}
\label{sec:ablation-text}
\paragraph{Estimator.}
Table~\ref{tab:est-ablation} ablates single factors against the \emph{Reference estimator}, an earlier variant with an extra context term ($17.038$\,dB $\text{PSNR}_{\text{all}}$, $24.6\%$ refresh). History only ($w \equiv 0$ under matched selection, $-0.961$\,dB; distinct from \emph{Eventful reuse} at $16.866$\,dB), in-flight only ($w \equiv 1$, $-0.652$\,dB), no affine correction ($W{=}I, b{=}0$: $-0.833$\,dB, $2.02$ steps/token), correlated-residual BLUE instead of the clipped gain ($-0.743$\,dB), and step-agnostic $(W,b)$ ($-0.276$\,dB) all reduce PSNR; disabling staleness tracking together with reuse-style selection gives $-0.463$\,dB. Removing norm restoration yields $+0.107$\,dB ($17.145$\,dB), so we do not claim it is necessary for 16-block PSNR. Removing the context term yields $+0.287$\,dB ($17.325$\,dB), which motivates the simplified estimator used throughout.

\paragraph{Refresh policy.}
Across 48-second policies (Tab.~\ref{tab:ablation}), step fusion ($14.92$\,dB) and risk ranking ($15.18$\,dB, $19.4\%$ flicker) each improve fidelity at a fixed $24.6\%$ budget. Readout-risk plans ($24.7\%$ refresh, $15.47$\,dB, \MainSteps{} steps/token) beat a KV-error proxy ($17.8\%$, $15.15$\,dB) in PSNR and early exit, at more refreshes and flicker ($23.5\%$). Triton reduces routing to \RoutingNew{}\,ms. A larger latency weight ($\lambda{=}2$) starves refresh to $10.9\%$ and lowers PSNR to $14.91$\,dB (Appendix Tab.~\ref{tab:screen}).

\section{Conclusion}
\label{sec:conclusion}
Once distillation and early exit make denoising cheap, the clean KV update dominates streaming video diffusion, and naive persistent-KV reuse merely shifts its cost into downstream denoising: \emph{cheap memory costs future compute}. \method breaks this loop on frozen causal video DiTs. It reconstructs most of the cache from two estimates already in GPU memory, through Pre-RoPE affine correction and clipped excess-error fusion, and reserves true $t=0$ refreshes for readout-critical tokens via tile-aligned plan selection and single-wave Triton execution. On 48-second LongLive-1.3B videos, it reduces DiT latency by \FullLatencyReduction{}\% (\FullSpeedup{}$\times$) and principal FLOPs by \FullTFLOPCut{}\% vs.\ dense KV updates, while lowering downstream steps below reuse ($\ReuseSteps{} \to \MainSteps{}$ steps/token) and improving Q4 PSNR over it by $+\PSNRGain{}$\,dB ($+\UntouchedGain{}$\,dB on held-out VidProM-16). On frozen Self-Forcing-DMD-1.3B, the same reconstruction curbs reuse-induced step inflation and reaches \SFDCFDynPSNR{}\,dB / \SFDCFDynVB{} \emph{Diag.} when stacked with chained KV updates and adaptive budgeting. These results suggest that persistent memory should be budgeted in closed loop with the denoising it conditions, rather than as an isolated per-block cost.

\paragraph{Limitations.}
The readout proxy is an offline first-order estimate, not an error bound; cross-backbone transfer needs a one-time training-free calibration to recover \emph{Diag.}; and we evaluate 1.3B DiTs on AMD GPUs without standard LPIPS, FVD, VBench-Long, or user studies (Appendix~\ref{sec:ext-limitations}).

\clearpage
\section*{LLM Usage Statement}
In this work we have not used LLMs to generate synthetic data sets, develop theoretical models or conceptual frameworks, formulate mathematical claims, provide critical ingredients for proving mathematical claims, assist in the writing of proofs, propose or refine hypotheses, or interpret results. We used LLMs to create and edit software code, such as experiment and evaluation scripts. We have reviewed all AI-assisted work. All LLM generated code was read, executed, and tested by the authors, and all reported results were checked against the saved experiment logs. Feedback from LLMs on methodology and experiments was treated as suggestions; the research questions, method design, and conclusions were determined by the authors. We take responsibility for the final content of this work, including text, claims or artifacts produced with the aid of LLMs.

{\small
\bibliographystyle{ieeenat_fullname}
\bibliography{references}
}
\clearpage
\maketitlesupplementary
\appendix
\setlength{\paraheadskip}{4pt}
\section{Implementation, Calibration, and Detailed Equations}
\label{sec:impl}

\begin{figure*}[!t]
\centering
\includegraphics[width=0.95\textwidth]{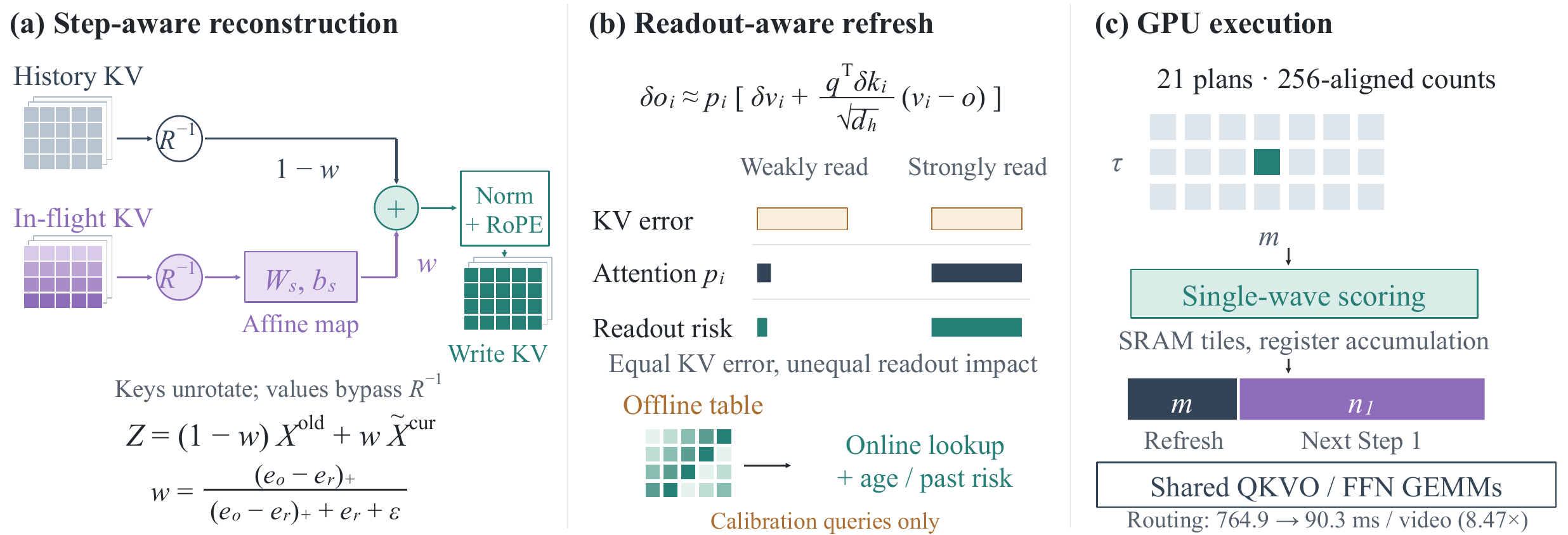}
\caption{\textbf{Detailed mechanisms of \method, complementing Fig.~\ref{fig:overview}.} \textbf{(a)} Keys are un-rotated into Pre-RoPE coordinates and values bypass rotation; in-flight KV receives the exit-step-stratified affine correction, is fused with history by the clipped excess-error weight $w$, and is written back after spherical norm restoration and destination 3D RoPE ($\approx 0.55\%$ of a chained clean-refresh token's FLOPs). Layer and head indices are omitted. \textbf{(b)} The first-order readout perturbation $\delta o_i$ weights KV error by its effect on attention output, suppressing unattended errors ($p_i \approx 0$) while emphasizing query-aligned disocclusions (bars are schematic). Risk is tabulated offline on calibration queries across 20 $(s_i, d_i)$ strata and looked up online with age and accumulated risk; note the KV-risk and readout-risk policies select different refresh rates ($17.8\%$ vs.\ $24.7\%$). \textbf{(c)} Single-wave scoring covers the $256$-aligned 21-plan lattice, and refresh rows share the QKVO/FFN GEMMs with the next block's Step~1; single-wave Triton routing cuts full-rollout routing from \RoutingOld{} to \RoutingNew{}\,ms (\RoutingSpeedup{}$\times$, $0.33\%$ DiT overhead).}
\label{fig:mechanisms}
\end{figure*}

\paragraph{Frozen generator architecture and memory layout.}
All primary experiments evaluate frozen LongLive-1.3B~\cite{yang2025longlive} causal weights built on Wan2.1-T2V-1.3B~\cite{wan2025}: $L=30$ transformer blocks, hidden width $D=1536$, FFN intermediate width $F=8960$, and $H=12$ self-attention heads of dimension $d_h=128$. Each latent video frame has shape $16\times60\times104$; $1\times2\times2$ spatiotemporal patching produces $30\times52=1{,}560$ tokens per frame and $N=4{,}680$ tokens per three-frame causal block ($64$-D raw latent patches, $0.57$\,MB/block in BF16). Text cross-attention context is padded to $T=512$ tokens with K/V projected once per rollout. LongLive's causal LoRA adapters are merged into BF16 weights prior to inference; \method performs zero weight fine-tuning or auxiliary network training. Persistent KV caches in the primary engine use a zero-copy U-shaped allocation across all compared arms: boundary spatial layers ($\ell \in \{0..5, 24..29\}$) maintain $12$ latent frames ($3$ sink + $9$ rolling frames, $18{,}720$ slots), whereas middle context-mixing layers ($\ell \in \{6..23\}$) maintain $7$ frames ($1$ sink + $6$ rolling frames, $10{,}920$ slots), bounding persistent KV memory at $2{,}468.0$\,MiB ($25.0\%$ below a uniform 12-frame cache of $3{,}290.6$\,MiB).

\paragraph{Closed-form ridge transport, gain rules, and norm restoration.}
In canonical Pre-RoPE coordinates (Fig.~\ref{fig:mechanisms}a), affine maps $(W_{\ell s h}, b_{\ell s h}) \in \R^{128\times128}\times\R^{128}$ (Eq.~\ref{eq:transport}) are fitted independently across $30$ layers, $4$ early-exit steps $s \in \{0,1,2,3\}$, and $24$ K/V heads using float64 sufficient moments $(\mu_x, \mu_y, C_{xx}, C_{xy})$ accumulated over steady-state blocks ($b \ge 2$) on eight development prompts (\emph{Dev-8}, $524{,}160$ steady-state tokens) via identity-centered ridge regression:
\begin{equation}
 W_{\ell s h} = (C_{xx} + \alpha I)^{-1}(C_{xy} + \alpha I), \quad
 b_{\ell s h} = \mu_y - \mu_x W_{\ell s h},
\label{eq:ridge}
\end{equation}
with $\alpha = 0.18\max(\operatorname{tr}(C_{xx})/128, 10^{-6})$ and $(W=I, b=0)$ fallback when $n < 512$. Across $30 \times 4 \times 24$ heads, $(W_{\ell s h}, b_{\ell s h})$ totals $47{,}554{,}560$ ($\approx 47.55$\,M) scalars ($90.7$\,MiB in BF16; this one-time offline pass was not timed separately).

Normalized correspondence distances $d_i$ are stratified into five bins via internal knots $\{0.12, 0.22, 0.34, 0.50\}$, forming $20$ $(s_i, d_i)$ strata. With Empirical Bayes shrinkage $n/(n+25)$ toward step-level marginals and cumulative-maximum monotonicity across distance bins, offline calibration compares diagonal inverse-variance weights $w_{\mathrm{ivw}} = e_o/(e_o + e_r)$, correlated-residual BLUE weights $w_{\mathrm{blue}} = \operatorname{clip}((e_o - c_{or})/(e_o + e_r - 2c_{or}), 0, 1)$, and clipped excess-error weights $w_{\mathrm{exc}} = (e_o - e_r)_+ / ((e_o - e_r)_+ + e_r + \varepsilon)$ (Eq.~\ref{eq:gain}). Source residuals exhibit $0.389$ mean correlation (up to $0.550$): diagonal $w_{\mathrm{ivw}}$ assigns $0.425$ mean weight on static $s_i=0, d_i < 0.12$ slots, and in closed loop the correlated $w_{\mathrm{blue}}$ variant trails the reference estimator, which uses $w_{\mathrm{exc}}$, by $0.743$\,dB $\text{PSNR}_{\text{all}}$ on HeldOut-16 (Tab.~\ref{tab:est-ablation-full}), whereas $w_{\mathrm{exc}}$ suppresses static $s_i=0$ weights to $0.036$ mean ($0.000$ in shallow layers) and rises to $0.686$ mean (up to $0.879$) on $(s_i=3, d_i > 0.50)$. Before re-applying destination 3D RoPE $R(p_{b,i})$ and BF16 writeback, gated slots apply spherical norm restoration to $Z_i$:
\begin{equation}
 \widehat{X}_i^{\mathrm{rec}} = Z_i \cdot \frac{(1 - w_{\ell c_i h})\norm{X_i^{\mathrm{old}}}_2 + w_{\ell c_i h}\norm{\widetilde{X}_i^{\mathrm{cur}}}_2}{\max(\norm{Z_i}_2, \varepsilon)}.
\label{eq:spherical}
\end{equation}
Norm restoration was introduced to prevent the Euclidean norm shrinkage observed in synthetic multi-round memory probes. Its effect on 16-block PSNR is small: removing it changes $\text{PSNR}_{\text{all}}$ by $+0.107$\,dB on HeldOut-16 (Tab.~\ref{tab:est-ablation-full}), so, as stated in \S\ref{sec:ablation-text}, we do not claim it is necessary at that horizon. The affine moments and the source-error tables are accumulated in the same offline \emph{Dev-8} pass, without cross-fitting.

\begin{figure}[!t]
\centering
\includegraphics[width=\linewidth]{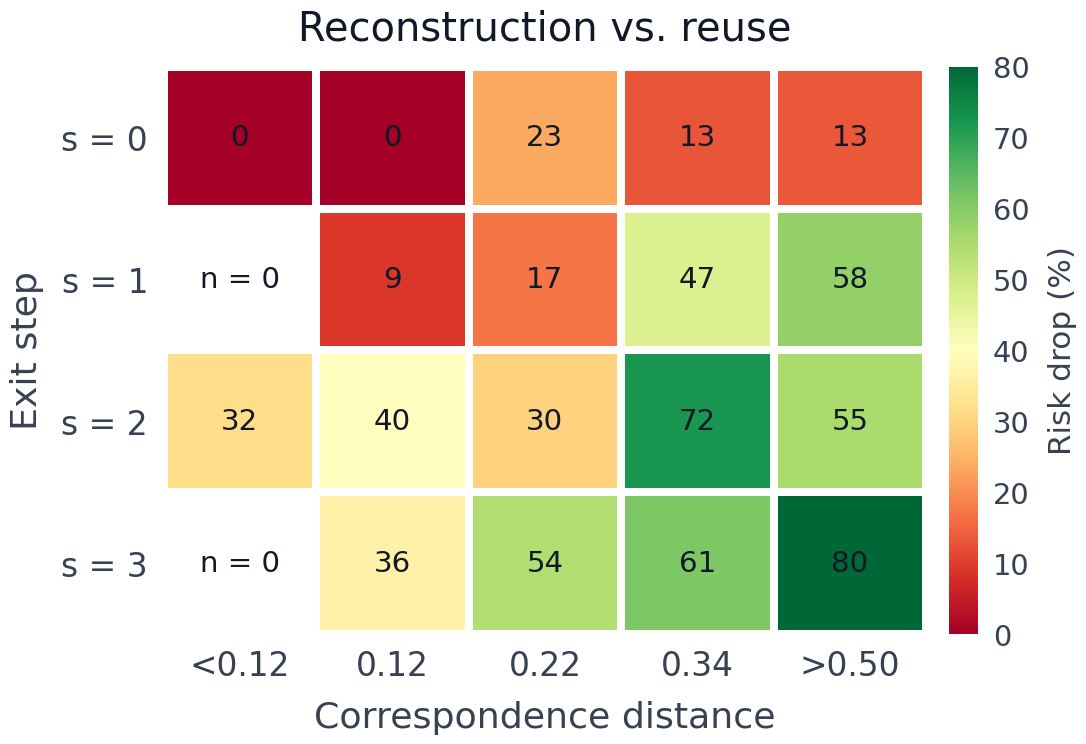}
\caption{\textbf{Offline readout-risk reduction from Pre-RoPE reconstruction across the $20$ $(s_i, d_i)$ calibration strata.} Cell values report percentage drop in offline normalized readout perturbation (Eq.~\ref{eq:risk}) relative to binary reuse, reaching $80\%$ on disoccluded late-exit tokens ($n{=}0$: empty calibration cell).}
\label{fig:risk}
\end{figure}

\paragraph{Offline readout tabulation and online heuristic equations.}
During offline risk calibration, our collector samples layers $\ell \in \{0, 5, 10, 15, 20, 25, 29\}$, token stride $73$ ($\approx 64$ slots/block), and next-block query stride $293$ ($16$ queries/block), replaying the deployed Triton fused assimilation-and-write kernel against dense clean-refresh reference KV on \emph{Dev-8}. Each of the $20$ $(s, d\text{-bin})$ cells accumulates normalized readout perturbation energy from Eq.~\eqref{eq:readout}:
\begin{equation}
 r_i(a) = \E_{\ell,h,q}\!\left[
 \frac{\norm{\delta o_i(a)}_2^2 / d_h}
 {\norm{o}_2^2 / d_h + \varepsilon}\right],
 \quad a \in \{\mathrm{reuse}, \mathrm{reconstruct}\},
\label{eq:risk}
\end{equation}
shrunk with Empirical Bayes pseudo-count $\kappa = 32$ toward global action means $\mu_a$ over cell counts $n_c$ and sums $S_c(a)$:
\begin{equation}
 \rho_c(a) = \frac{S_c(a) + \kappa\mu_a}{n_c + \kappa} \cdot
 \frac{1}{\max(\mu_{\mathrm{reuse}}, \varepsilon)},
\label{eq:normalized-risk}
\end{equation}
The resulting tables (Fig.~\ref{fig:risk}; fitting seed $830021$) agree with a separate \emph{Dev-8} validation seed ($930029$) within $0.020$ normalized KV MAE and $2.4\times10^{-5}$ raw readout MAE. At online block $b$ (Alg.~\ref{alg:samr} in \S\ref{sec:method}), routing looks up $\rho_{c_i}(a)$ without observing live future queries and applies the empirical age/risk recurrence ($u_i = u_{\mathrm{prev}}[\pi(i)]$, $h_i = h_{\mathrm{prev}}[\pi(i)] + 1$):
\begin{equation}
 \widehat{r}_i(a) = \rho_{c_i}(a)\bigl(1 + \gamma_h \min(h_i, H_{\max})\bigr) + \gamma_u^{(a)} u_i,
\label{eq:lineage}
\end{equation}
with $\gamma_h = 0.05$, $H_{\max} = 12$, $\gamma_u^{(\mathrm{reuse})} = 0.25$, and $\gamma_u^{(\mathrm{recon})} = 0.10$ (reverting $\widehat{r}_i(\mathrm{reconstruct}) \leftarrow \widehat{r}_i(\mathrm{reuse})$ when $d_i < \tau_d = 0.22$ and $s_i < \tau_s = 2$). With $r_{\min,i} = \min(\widehat{r}_i(\mathrm{reuse}), \widehat{r}_i(\mathrm{reconstruct}))$, clean refresh uses residual floor $\widehat{r}_{\mathrm{refresh},i} = \eta\,r_{\min,i}$ ($\eta = 0.1$), tokens are ordered by $\psi_i = (1-\eta)\,r_{\min,i}(1 + \operatorname{clip}(d_i, 0, 2))$, and gain cutoffs $\tau \in \mathcal{C} = \{0, 0.5, 1.0\}$ threshold against $\bar{g}_b = \frac{1}{N}\sum_{j=1}^N (\widehat{r}_j(\mathrm{reuse}) - \widehat{r}_j(\mathrm{reconstruct}))_+$ across the $\Delta_m = 256$-aligned refresh counts $\mathcal{M} = \{512, 768, 1024, 1280, 1536, 1792, 2304\}$ ($|\mathcal{P}_b| = 21$, $T_0 = 100$\,ms, $\lambda = 1.0$).

\paragraph{Hardware latency model and 8-GPU kernel microbenchmarks.}
The NNLS hardware cost model predicts merged KV-update + Step-1 duration (ms) from nine nonnegative features: intercept ($204.31$\,ms), refresh fraction $m/N$ ($119.58$\,ms), four exit-step reconstruction fractions ($22.42, 27.35, 28.44, 25.13$\,ms for $s=0..3$), fraction of nonempty exit-step groups ($2.69$\,ms), initial cache ramp $\min(b,3)/3$ ($4.10$\,ms), and next-Step-1 indicator ($576$ profiling runs; fit MAE $0.99$\,ms / $0.39\%$; held-out repeat MAE $0.97$\,ms / $0.38\%$; $2{,}790$-block test MAE \CostModelMAE{}\,ms / \CostModelRelErr{}\%, Pearson $r=0.943$; Fig.~\ref{fig:dividend}c). Across all eight AMD GPUs ($62$ steady-state calls/video), Triton $5\times5$ matching runs in $5.42$\,ms mean \vs $146.09$\,ms in PyTorch ($27.0\times$ 8-GPU mean; \CorrSpeedup{}$\times$ on GPU0), and single-wave 21-plan SRAM scoring runs in $48.99$\,ms mean \vs $567.33$\,ms ($11.6\times$ 8-GPU mean; \PlanSpeedup{}$\times$ on GPU0, $\max|\Delta| = 0$), cutting full-video routing from \RoutingOld{} to \RoutingNew{}\,ms (\RoutingSpeedup{}$\times$, $0.33\%$ DiT overhead).

\section{Exact Arithmetic Accounting and Closed-Loop Cost}
\label{sec:accounting}
\begin{figure*}[!t]
\centering
\includegraphics[width=0.95\textwidth]{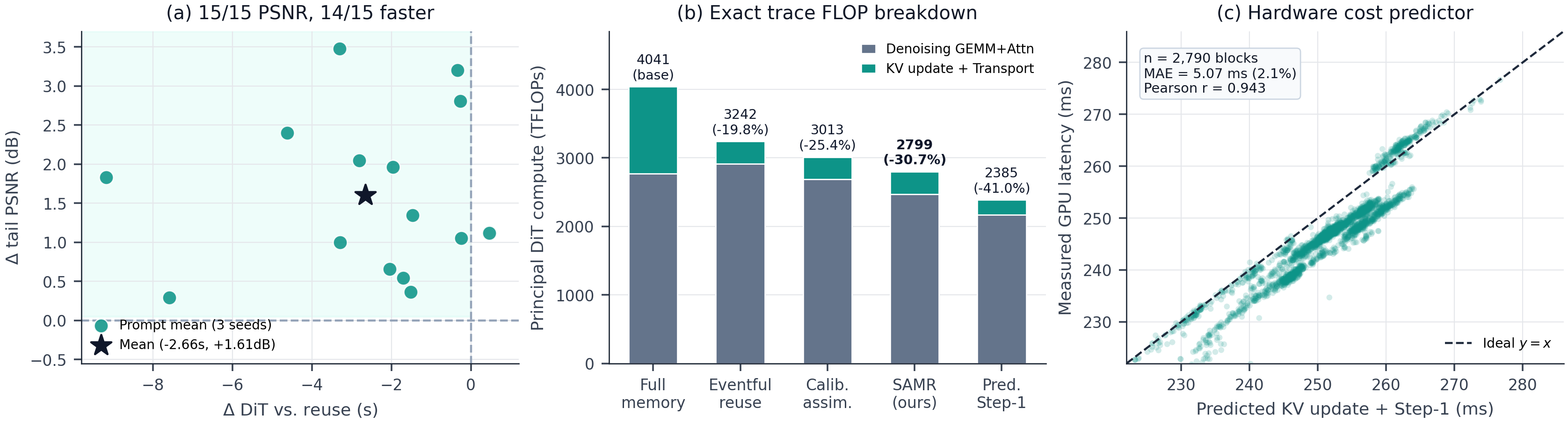}
\caption{\textbf{Per-prompt paired deltas, trace-exact TFLOP breakdown, and NNLS cost-predictor accuracy (\texttt{Main 15$\times$3}).} \textbf{(a)} Three-seed prompt-mean deltas (\method minus Eventful reuse): $15/15$ prompts improve Q4 PSNR and $14/15$ run faster. \textbf{(b)} Trace-exact principal DiT TFLOPs ($-\FullTFLOPCut{}\%$ vs.\ dense; exploratory \emph{Pred.\ Step-1} shown on right). \textbf{(c)} NNLS KV-update + Step-1 latency predictor vs.\ measured GPU time across $2{,}790$ blocks (\CostModelMAE{}\,ms MAE, \CostModelRelErr{}\% error, $r=0.943$).}
\label{fig:dividend}
\end{figure*}

Executing $q$ active query rows over $k$ visible KV slots in one transformer layer requires principal FLOPs ($1\text{ MAC}=2\text{ FLOPs}$):
\begin{equation}
 \mathcal{F}(q, k) = q \Bigl[ 12D^2 + 4DF + 4D(k + T) \Bigr],
\label{eq:flops}
\end{equation}
and closed-loop streaming couples write-side KV updates with downstream early-exit denoising across blocks:
\begin{equation}
 \mathcal{C}_{\mathrm{tot}}(M_{\le B}) = \sum_{b=0}^{B-1} \Bigl[ \mathcal{C}_{\mathrm{den}}(x_b \mid M_{<b}) + \mathcal{C}_{\mathrm{upd}}(M_b) \Bigr].
\label{eq:closed-loop-cost}
\end{equation}
With hardware-aligned active denoising rows $S_b = \sum_{s=0}^3 A_{b,s}$, refreshed KV-update rows $M_b$, and reconstructed slots $U_b$, principal 64-block DiT FLOPs total $\sum_{b=0}^{63}\sum_{\ell=0}^{29}[\mathcal{F}(S_b, K_{b,\ell}) + \mathcal{F}(M_b, J_{b,\ell})] + \sum_{b=1}^{62} 4HL U_b d_h^2$, where $K_{b,\ell} = \min((b+1)N, C_\ell)$ ($C_\ell \in \{10{,}920, 18{,}720\}$) and $J_{b,\ell} = 2N = 9{,}360$ for $b \ge 1$ (omitting the unconsumed KV update at block $63$ uniformly across all arms: $\FullTFLOPs{}$, $\ReuseTFLOPs{}$, and $\MainTFLOPs{}$\,TFLOPs for \emph{Full memory}, \emph{Eventful reuse}, and \method; Fig.~\ref{fig:dividend}b). Reconstructing $U_b$ gated slots adds $23.59\,\mathrm{MFLOPs}\times U_b$ (\MainReconTFLOPs{}\,TFLOPs per video in \method). Timing logs record merged KV-update + next-block Step-1 (\FullCommitStepOneS{}, \ReuseCommitStepOneS{}, \MainCommitStepOneS{}\,s) separately from remaining Steps~2--4 (\FullDenoiseStageS{}, \ReuseDenoiseStageS{}, \MainDenoiseStageS{}\,s) and routing.

\section{Diagnostic Metric and Statistical Definitions}
\label{sec:metrics}
\paragraph{Reference-trajectory agreement.}
Given candidate RGB $I_b \in [0,1]^{T_b \times 3 \times H \times W}$ and same-seed \emph{Full memory} reference $I_b^*$, per-block PSNR is $-10\log_{10}\operatorname{mean}((I_b - I_b^*)^2)$, averaged over 12-second quarters (Q1: blocks $1$--$15$; Q2: $16$--$31$; Q3: $32$--$47$; Q4: $48$--$63$). For block mean saturation ($\max_c I - \min_c I$), motion ($\operatorname{mean}|I_{t+1} - I_t|$), or second-order flicker ($\operatorname{mean}|I_{t+2} - 2I_{t+1} + I_t|$), percentage drift is $d_g(b) = 100 |g(I_b)/\max(g(I_b^*), \varepsilon) - 1|$. The tables that report SSIM and \emph{AlexD.} (uniform-cache and extended-stack experiments) evaluate post-anchor frames ($t \ge 12$) via 12-frame mean SSIM~\cite{wang2004ssim} and 16-frame \emph{normalized AlexNet distance} (\emph{AlexD.}): across five ImageNet-pretrained AlexNet slices (\texttt{relu1}--\texttt{relu5}, without learned LPIPS~\cite{zhang2018lpips} linear channel weights), unit-channel-normalized features $\hat{\phi}_k = \phi_k / (\norm{\phi_k}_2 + 10^{-8})$ are averaged as $\frac{1}{5}\sum_{k=1}^5 \operatorname{mean}_{h,w}\norm{\hat{\phi}_k(I) - \hat{\phi}_k(I^*)}_2^2$.

\paragraph{No-reference diagnostics and bootstrap protocol.}
CLIP ViT-B/32 (\texttt{laion2b\_s34b\_b79k})~\cite{radford2021clip,schuhmann2022laion5b} measures cosine similarity to early anchor frames (\emph{Anchor}) and prompt text ($c_{\mathrm{txt}}$). Our custom diagnostic composite (\emph{Diag.}) combines five VBench-inspired~\cite{huang2024vbench} terms:
\begin{align}
 \mathrm{Diag} &= 0.25\,S_{\mathrm{subj}} + 0.25\,S_{\mathrm{smooth}} + 0.20\,S_{\mathrm{flick}} \nonumber\\
 &\quad + 0.15\,(250\,c_{\mathrm{txt}}) \nonumber\\
 &\quad + 0.15\max\bigl(0,\, 100 - 0.5\,d_{\mathrm{sat}}^{\mathrm{self}} - 0.5\,d_{\mathrm{sharp}}^{\mathrm{self}}\bigr),
\label{eq:diagcomp}
\end{align}
where $S_{\mathrm{subj}} = 100(0.5\,\overline{\mathrm{sim}}(f_t, f_{t-1}) + 0.5\,\overline{\mathrm{sim}}(f_t, \bar{f}_{0:1}))$, Farneback~\cite{farneback2003flow} optical flow at $240\times416$ defines static stability $S_{\mathrm{flick}} = 100(1 - \mathrm{MAE}_{\mathrm{static}})$ (flow $<1$\,px) and midpoint warp smoothness $S_{\mathrm{smooth}} = 100(1 - \mathrm{MAE}_{\mathrm{interp}})$, and $d_{\mathrm{sat}}^{\mathrm{self}}, d_{\mathrm{sharp}}^{\mathrm{self}}$ measure first-to-last quarter self-drift. On \texttt{Main 15$\times$3} ($15$ prompts $\times 3$ seeds), post-hoc paired $95\%$ confidence intervals average the $3$ seed deltas per prompt first, then draw $10{,}000$ percentile-bootstrap resamples of the $15$ prompt means with a fixed random seed. Table~\ref{tab:paired} lists the resulting intervals for \method against \emph{Eventful reuse}: the DiT-time and Q4 PSNR intervals exclude zero, whereas those of the three drift metrics span it. The \emph{Dev-8} correlation quoted in \S\ref{sec:preliminaries} is the Spearman rank correlation between the per-token difference of the Step-1 and Step-4 clean predictions and the between-block patch change, computed per probed block: mean $0.643$, standard deviation $0.162$ across blocks.

\begin{table}[h]
\centering
\small
\setlength{\tabcolsep}{5pt}
\begin{tabular}{@{}lrcc@{}}
\toprule
Metric (\method $-$ reuse) & Mean & $95\%$ interval & Better on \\
\midrule
DiT time (s) & $\MainDitDelta{}$ & $[\MainDitCILo{},\, \MainDitCIHi{}]$ & $14/15$ \\
Q4 PSNR (dB) & $+\PSNRGain{}$ & $[\MainPSNRCILo{},\, \MainPSNRCIHi{}]$ & $15/15$ \\
Q4 saturation drift (\%) & $\MainSatDelta{}$ & $[\MainSatCILo{},\, \MainSatCIHi{}]$ & -- \\
Q4 flicker drift (\%) & $\MainFlkDelta{}$ & $[\MainFlkCILo{},\, \MainFlkCIHi{}]$ & -- \\
Q4 motion drift (\%) & $\MainMotDelta{}$ & $[\MainMotCILo{},\, \MainMotCIHi{}]$ & -- \\
\bottomrule
\end{tabular}
\caption{\textbf{Paired prompt-level statistics on \texttt{Main 15$\times$3}} (\method minus \emph{Eventful reuse}; $15$ prompt means, $10{,}000$ percentile-bootstrap draws). \emph{Better on}: number of prompts on which \method improves over reuse; not tabulated for the drift metrics, whose intervals span zero.}
\label{tab:paired}
\end{table}

\section{Optional System Extensions: VAE Delivery, Low-Bit Scaling, and Waterfall}
\label{sec:delivery}
\label{sec:fullstack}
The results in this section are optional extensions and lie outside the claims of the main paper. The delivery study keeps the \method DiT of \S\ref{sec:experiments} fixed and changes only how VAE decoding is scheduled; the low-bit study uses a separate extended-stack configuration that adds exploratory components.

\begin{figure*}[!t]
\centering
\includegraphics[width=0.95\textwidth]{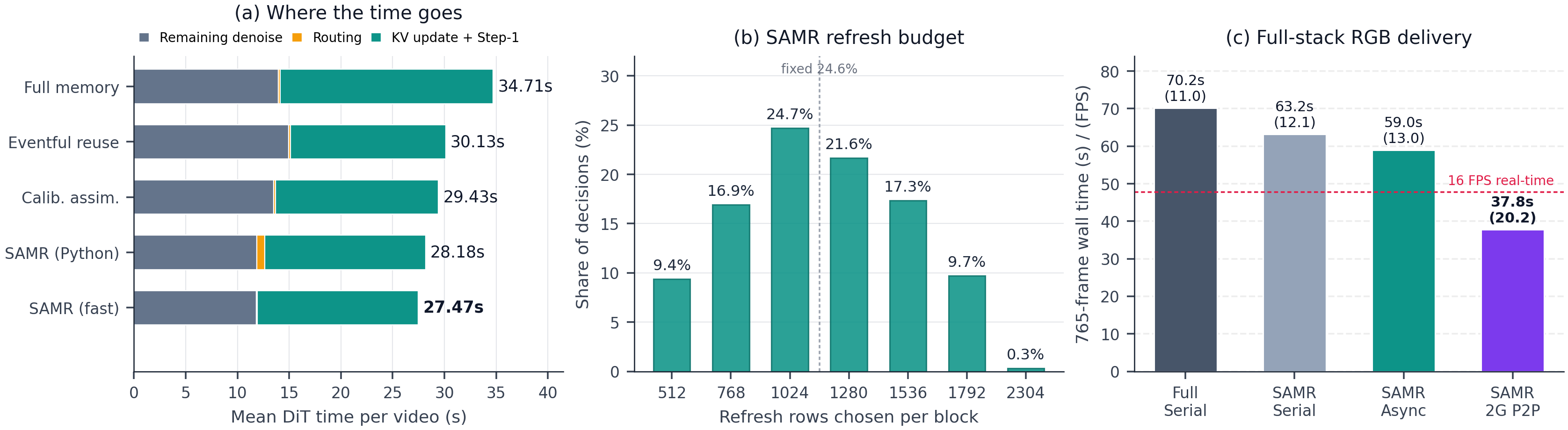}
\caption{\textbf{Stage-wise GPU wall time, adaptive tile selection, and end-to-end RGB delivery.} \textbf{(a)} Per-video DiT stage breakdown (\method: KV update + Step-1 $\FullCommitStepOneS{} \to \MainCommitStepOneS{}$\,s, remaining denoise $\ReuseDenoiseStageS{} \to \MainDenoiseStageS{}$\,s vs.\ reuse, routing $\RoutingOld{} \to \RoutingNew{}$\,ms). \textbf{(b)} Selected refresh tiles $m^*$ across $2{,}790$ steady-state blocks. \textbf{(c)} 765-frame RGB delivery across serial, 1-GPU async, and 2-GPU P2P.}
\label{fig:system}
\end{figure*}

\paragraph{Narrow-waist causal VAE pipelining.}
With 64-block DiT execution at \MainLatency{}\,s, causal 3D VAE decoding ($\sim$35.2--35.8\,s for 765 frames) accounts for $56.6\%$ of \method's serial RGB delivery (\DelivSerial{}\,s). Finalized BF16 block latents $x_b$ ($1\times3\times16\times60\times104$, $0.57$\,MB/block vs.\ $14.4$\,MB FP16 RGB) are ready before $\mathcal{U}_\theta$ updates $M_b$. Across eight 765-frame seed-0 clips (Tab.~\ref{tab:delivery}, Fig.~\ref{fig:system}c; host timestamps are taken after the device-to-host copy completes), with the \method generator held fixed, \emph{1-GPU async SDMA} lowers wall time from \DelivSerial{} to \DelivAsync{}\,s ($-\DelivAsyncCut{}\%$), trading higher first-block latency ($\DelivSerialFirst{} \to \DelivAsyncFirst{}$\,s) for higher throughput (\DelivAsyncFPS{}\,FPS; $-\DelivAsyncVsFullGPUCut{}\%$ vs.\ \emph{Full serial}'s \DelivFullSerial{}\,s). Offloading VAE decode via \emph{2-GPU P2P} reaches \DelivTwoGPU{}\,s (\DelivTwoGPUFPS{}\,FPS) at higher total GPU reservation (\DelivTwoGPUSec{} \vs \DelivAsyncGPU{}\,GPU$\cdot$s). Both modes are bit-exact ($\max|\Delta\mathrm{RGB}|=0$) relative to serial VAE decoding of the same \method generator. These are finite-clip averages: a mean above $16$\,FPS does not by itself guarantee stall-free unbounded serving, which additionally requires bounded-queue backpressure.

\begin{table}[!t]
\centering
\small
\setlength{\tabcolsep}{1.5pt}
\begin{tabular}{@{}lrrrrrrr@{}}
\toprule
Pipeline & GPU & First$\downarrow$ & Gap$\downarrow$ & Wall$\downarrow$ & Spd.$\uparrow$ & GPU$\cdot$s & FPS$\uparrow$ \\
 & & (s) & (s) & (s) & vs.\ Full & & \\
\midrule
Full (serial) & 1 & -- & -- & 70.22 & 1.00$\times$ & 70.2 & 10.96 \\
Reuse (serial) & 1 & -- & -- & 66.18 & 1.06$\times$ & 66.2 & 11.64 \\
\midrule
SAMR serial & 1 & \textbf{0.903} & 0.982 & 63.23 & 1.11$\times$ & 63.2 & 12.14 \\
SAMR async & 1 & 1.256 & 0.909 & 58.98 & 1.19$\times$ & \textbf{59.0} & 13.03 \\
\textbf{SAMR P2P} & 2 & 1.096 & \textbf{0.576} & \textbf{37.79} & \textbf{1.86$\times$} & 75.6 & \textbf{20.25} \\
\bottomrule
\end{tabular}

\caption{\textbf{Ordered 765-frame RGB delivery ($8$ clips, $480\times832$).} Async and 2-GPU P2P are bit-exact ($\max|\Delta\mathrm{RGB}|=0$) vs.\ serial decoding of the \emph{same} \method generator.}
\label{tab:delivery}
\end{table}

\begin{table}[!t]
\centering
\small
\setlength{\tabcolsep}{1.15pt}
\begin{tabular}{@{}llrrrrr@{}}
\toprule
Precision & Role & PSNR$\uparrow$ & SSIM$\uparrow$ & AlexD.$\downarrow$ & Flk.$\downarrow$ & Diag.$\uparrow$ \\
 & (quant.) & (dB) & & (norm.) & (\%) & (comp.) \\
\midrule
FP8 (\texttt{e4m3}) & KV update & \textbf{19.14} & \textbf{0.717} & \textbf{0.427} & \textbf{7.7} & 94.10 \\
 & Denoise & 15.57 & 0.560 & 0.749 & 10.8 & 93.90 \\
\midrule
W4A8 (g128) & KV update & \textbf{16.10} & \textbf{0.596} & \textbf{0.656} & \textbf{19.3} & 93.68 \\
 & Denoise & 12.14 & 0.396 & 1.175 & 23.5 & 93.74 \\
\midrule
W4A4 (g128) & KV update & \textbf{14.93} & \textbf{0.549} & \textbf{0.775} & \textbf{23.8} & 93.40 \\
 & Denoise & 11.54 & 0.333 & 1.271 & 80.5 & 92.23 \\
\bottomrule
\end{tabular}

\caption{\textbf{Role-asymmetric quantization sensitivity on HeldOut-16 (189 frames, extended-stack configuration), complementing Fig.~\ref{fig:fullstack}b.} Quantizing only $\mathcal{U}_\theta$ (KV update) while keeping $\mathcal{D}_\theta$ in BF16 preserves $+3.4$ to $+4.0$\,dB higher PSNR than quantizing $\mathcal{D}_\theta$.}
\label{tab:rolequant}
\end{table}

\begin{figure*}[!t]
\centering
\includegraphics[width=0.93\textwidth]{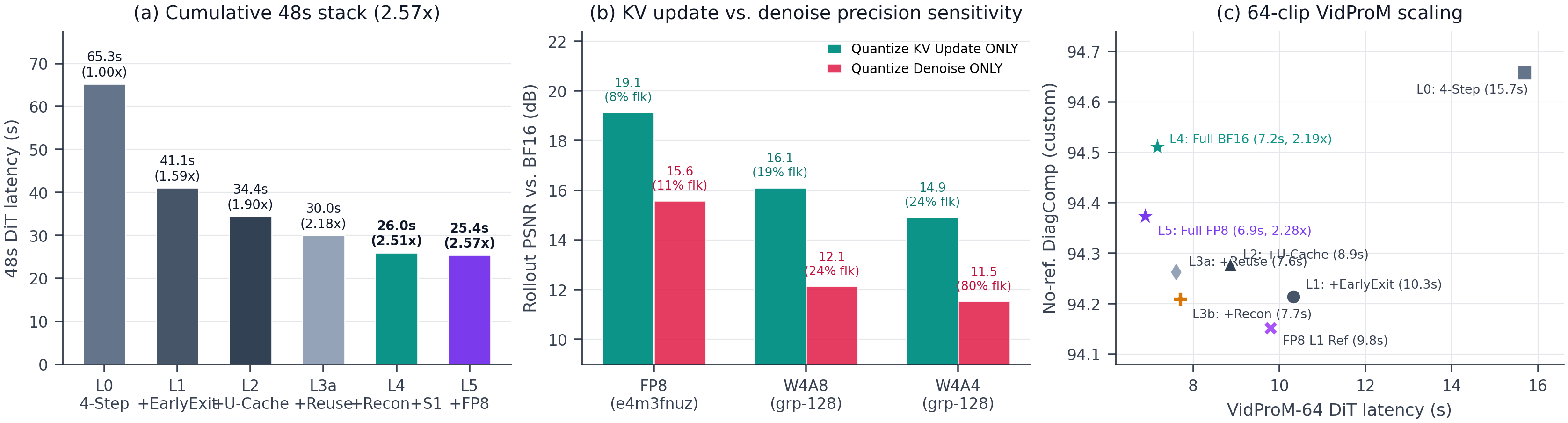}
\caption{\textbf{Extended cumulative system waterfall, quantization role asymmetry, and 64-clip VidProM-64 scaling.} \textbf{(a)} 48-second L0$\to$L5 progression from 4-step dense BF16 without early exit (\WaterfallBaseLong{}\,s) to full-stack + Predictive Step-1 + FP8 (\WaterfallFastLong{}\,s, \WaterfallSpdLong{}$\times$; \emph{Diag.} $92.95 \to 93.60$). \textbf{(b)} Role-isolated KV-update vs.\ denoise quantization. \textbf{(c)} 64-clip \emph{VidProM-64} short-horizon scaling (\WaterfallSpdWide{}$\times$). L1--L2 are existing techniques that the main-paper baseline already includes (Tab.~\ref{tab:waterfall}).}
\label{fig:fullstack}
\end{figure*}

\begin{table*}[!t]
\centering
\small
\setlength{\tabcolsep}{2.2pt}
\begin{tabular}{@{}lrrrrrrrrrrrr@{}}
\toprule
 & \multicolumn{3}{c}{HeldOut-16 (189f)} & \multicolumn{3}{c}{VidProM-64 (64 clips)} & \multicolumn{3}{c}{Long-64 (765f / 48\,s)} & \multicolumn{3}{c}{SFD-1.3B (16 clips)} \\
\cmidrule(lr){2-4}\cmidrule(lr){5-7}\cmidrule(lr){8-10}\cmidrule(lr){11-13}
Optimization Stage & DiT$\downarrow$ & Spd.$\uparrow$ & Diag.$\uparrow$ & DiT$\downarrow$ & Spd.$\uparrow$ & Diag.$\uparrow$ & DiT$\downarrow$ & Spd.$\uparrow$ & Diag.$\uparrow$ & DiT$\downarrow$ & Spd.$\uparrow$ & Diag.$\uparrow$ \\
\midrule
L0: 4-step dense DiT (BF16) & 15.68 & 1.00$\times$ & 93.36 & 15.69 & 1.00$\times$ & 94.66 & 65.30 & 1.00$\times$ & 92.95 & 15.70 & 1.00$\times$ & 91.86 \\
L1: + Speculative token early exit & 10.13 & 1.55$\times$ & 93.90 & 10.33 & 1.52$\times$ & 94.21 & 41.12 & 1.59$\times$ & 93.76 & 11.30 & 1.39$\times$ & 92.13 \\
L2: + U-shaped + chained KV & 8.72 & 1.80$\times$ & 94.30 & 8.87 & 1.77$\times$ & 94.28 & 34.42 & 1.90$\times$ & 93.43 & -- & -- & -- \\
\midrule
L3a: + Eventful reuse ($24.6\%$) & 7.68 & 2.04$\times$ & 93.40 & 7.61 & 2.06$\times$ & 94.26 & 29.99 & 2.18$\times$ & 92.37 & 8.64 & 1.82$\times$ & 92.19 \\
L3b: + Pre-RoPE recon. ($24.6\%$) & 7.65 & 2.05$\times$ & 93.90 & 7.70 & 2.04$\times$ & 94.21 & 29.37 & 2.22$\times$ & 93.12 & -- & -- & -- \\
L4: + Adaptive budget + Pred. Step-1 (BF16) & 6.99 & 2.24$\times$ & 94.05 & 7.17 & 2.19$\times$ & 94.51 & 26.01 & 2.51$\times$ & 93.51 & 7.97 & 1.97$\times$ & 91.97 \\
\midrule
FP8 ref.: L1 full memory (\texttt{e4m3fnuz}) & 9.61 & 1.63$\times$ & 93.90 & 9.79 & 1.60$\times$ & 94.15 & 39.05 & 1.67$\times$ & 93.68 & 10.46 & 1.50$\times$ & 92.28 \\
\textbf{L5: Full-stack + FP8 (\texttt{e4m3fnuz})} & \textbf{6.80} & \textbf{2.31$\times$} & 94.05 & \textbf{6.89} & \textbf{2.28$\times$} & 94.37 & \textbf{25.44} & \textbf{2.57$\times$} & 93.60 & \textbf{7.36} & \textbf{2.13$\times$} & 92.22 \\
\bottomrule
\end{tabular}

\caption{\textbf{Cumulative L0$\to$L5 extended system waterfall across four suites} (HeldOut-16, 64-clip VidProM-64, 48-second Long-64, and Self-Forcing-DMD-1.3B) in the extended-stack configuration. Speedups are relative to L0, 4-step dense BF16 denoising \emph{without} early exit, within each suite. L1--L2 are existing techniques and already give $1.90\times$ at 48\,s; the main-paper engine, and hence its \emph{Full memory} baseline, includes them (\S\ref{sec:setup}), so these cumulative factors are not comparable to the $\FullSpeedup{}\times$ of Tab.~\ref{tab:main}. L4--L5 add exploratory \emph{Predictive Step-1} pruning, which is not part of \method.}
\label{tab:waterfall}
\end{table*}

\paragraph{Cumulative low-bit stack and role asymmetry.}
In the separate extended-stack configuration (Fig.~\ref{fig:fullstack}, Tabs.~\ref{tab:rolequant} and~\ref{tab:waterfall}), combining prior early-exit (L1) and U-cache / chained KV-update optimizations (L2) with Pre-RoPE reconstruction, exploratory \emph{Predictive Step-1} token pruning (L4), and AMD hardware FP8 (\texttt{e4m3fnuz}, L5) lowers 48-second DiT runtime from uncompressed 4-step BF16 (\WaterfallBaseLong{}\,s) to \WaterfallFastLong{}\,s (\WaterfallSpdLong{}$\times$; \WaterfallSpdWide{}$\times$ on 12-s \emph{VidProM-64} and \WaterfallSpdSFD{}$\times$ on SFD-1.3B). Of the $\WaterfallSpdLong{}\times$, $1.90\times$ comes from L1--L2, which the main-paper baseline already includes. \emph{Predictive Step-1} remains exploratory: the \texttt{Main 15$\times$3} variant that includes it is faster ($25.01$\,s) but has lower Q4 PSNR ($14.35$ vs.\ $15.47$\,dB) and higher flicker drift ($27.25\%$ vs.\ $23.46\%$) than \method (Tab.~\ref{tab:exploratory}), a cost that \emph{Diag.} does not register; we therefore exclude it from \method. Role-isolated precision tests (Tab.~\ref{tab:rolequant}) confirm that quantizing only $\mathcal{U}_\theta$ retains $+3.4$ to $+4.0$\,dB higher rollout PSNR than quantizing $\mathcal{D}_\theta$ across FP8, W4A8, and W4A4.

\section{Configurations, Prompt Splits, and Full Tables}
\label{sec:full-tables}

\begin{table*}[!t]
\centering
\begin{minipage}[t]{0.485\textwidth}
\centering
\footnotesize
\setlength{\tabcolsep}{2.2pt}
\resizebox{\linewidth}{!}{%
\begin{tabular}{@{}lll@{}}
\toprule
Evidence & Configuration & Role \\
\midrule
$\FullLatency{}{\to}\MainLatency{}$\,s (\FullSpeedup{}$\times$) & Primary engine (BF16) & Main (Tab.~\ref{tab:main}) \\
Tabs.~\ref{tab:quality-transfer}, \ref{tab:est-ablation} & Uniform 12-frame cache & Mechanism \& abl. \\
$\WaterfallBaseLong{}{\to}\WaterfallFastLong{}$\,s (\WaterfallSpdLong{}$\times$) & Ext.\ stack + Pred.\ S1 + FP8 & Optional stack \\
SFD-1.3B zero-shot & Fixed Pre-RoPE (\llcal{}) & Component transfer \\
$\DelivTwoGPU{}$\,s, $\DelivTwoGPUFPS{}$\,FPS & Primary engine + 2-GPU VAE & Optional delivery \\
\bottomrule
\end{tabular}}
\caption{\textbf{Configuration attribution across main and supplementary results.} The headline claims use the BF16 primary engine of \S\ref{sec:setup}; the two optional rows are outside the main claims.}
\label{tab:config-map}
\end{minipage}\hfill
\begin{minipage}[t]{0.495\textwidth}
\centering
\footnotesize
\setlength{\tabcolsep}{1.7pt}
\begin{tabular}{@{}llll@{}}
\toprule
Split & Seeds / Horizon & Role & Prior use \\
\midrule
Dev-8 ($8$) & $55\text{k}{+},830\text{k},730\text{k}$ / 12,48s & Fit/screen & Development \\
Main $15{\times}3$ & $130\text{k}$--$330\text{k}$ / 48s (12s) & Main/abl. & Reused in dev. \\
VidProM-64 & $66\text{k}{+}100i$ / 12s (189f) & Short ext. & 12-s runs, no fit \\
VidProM-16 & $880021$ (new) / 48s (765f) & Locked 48s & Texts: VidProM-64 \\
\bottomrule
\end{tabular}
\caption{\textbf{Prompt-split provenance.} \emph{Prior use}: how a split was used before the reported runs. Because \texttt{Main 15$\times$3} was evaluated repeatedly during development, we also report \texttt{VidProM-16}: a fresh seed and $4\times$ longer horizon under the locked policy, without fitting or screening, on 16 prompt texts from the 12-s \texttt{VidProM-64} pool.}
\label{tab:prompt-history}
\end{minipage}
\end{table*}

\begin{table*}[!t]
\centering
\small
\setlength{\tabcolsep}{2.7pt}
\begin{tabular}{@{}lrrrrrrrrrrrr@{}}
\toprule
 & \multicolumn{6}{c}{DiT Execution, Compute \& Memory Actions} & \multicolumn{4}{c}{Ref.-Trajectory Agreement (Q4)} & \multicolumn{2}{c}{No-Ref.\ CLIP} \\
\cmidrule(lr){2-7}\cmidrule(lr){8-11}\cmidrule(lr){12-13}
Method & DiT $\downarrow$ & Spd. $\uparrow$ & TFLOPs $\downarrow$ & Steps $\downarrow$ & Ref. & Rec. & PSNR $\uparrow$ & Sat. $\downarrow$ & Flk. $\downarrow$ & Mot. $\downarrow$ & Anc. $\uparrow$ & Text $\uparrow$ \\
 & (s) & vs.\ Full & /vid & /tok & (\%) & (\%) & (dB) & (\%) & (\%) & (\%) & (CLIP) & (CLIP) \\
\midrule
\textbf{SAMR (ours)} & \textbf{27.47} & \textbf{1.26$\times$} & \textbf{2799} & \textbf{1.58} & 24.7 & 45.2 & \textbf{15.47} & 8.70 & 23.46 & 22.81 & 0.865 & \textbf{0.340} \\
\midrule
Recalibrated / KV & 28.63 & 1.21$\times$ & 2944 & 1.71 & 20.9 & 41.4 & 14.75 & 8.16 & 21.46 & 19.57 & 0.866 & 0.336 \\
Recalibrated / readout & 28.07 & 1.24$\times$ & 2868 & 1.68 & 17.6 & 43.6 & 14.50 & 8.71 & 22.30 & 20.71 & 0.866 & 0.339 \\
Predictive Step-1 variant & 25.01 & 1.39$\times$ & 2385 & 1.39 & 15.5 & 36.1 & 14.35 & 12.30 & 27.25 & 27.45 & 0.869 & 0.337 \\
\bottomrule
\end{tabular}

\caption{\textbf{Exploratory multi-factor variants on \texttt{Main 15$\times$3} (45 rollouts)}, compared against \method. \emph{Recalibrated KV / readout} bundle per-layer normalization, isotonic risk tables, anchor-drift penalties, and empirical refresh curves; \emph{Predictive Step-1} additionally prunes stationary Step-1 tokens. None of these variants is part of \method: each lowers Q4 PSNR relative to it.}
\label{tab:exploratory}
\end{table*}

\begin{table*}[!t]
\centering
\small
\setlength{\tabcolsep}{2.2pt}
\begin{tabular}{@{}lrrrrrrrrrr@{}}
\toprule
 & \multicolumn{3}{c}{Execution} & \multicolumn{3}{c}{No-Ref.\ Quality Diagnostics} & \multicolumn{4}{c}{Ref.-Trajectory Agreement (12f Dense)} \\
\cmidrule(lr){2-4}\cmidrule(lr){5-7}\cmidrule(lr){8-11}
Method & DiT $\downarrow$ & Steps $\downarrow$ & Ref. & Diag. $\uparrow$ & Subj. $\uparrow$ & Smooth. $\uparrow$ & SSIM $\uparrow$ & AlexD. $\downarrow$ & PSNR $\uparrow$ & Sat. $\downarrow$ \\
 & (s) & /tok & (\%) & (comp.) & (\%) & (\%) & & (norm.) & (dB) & (\%) \\
\midrule
Full memory (12-frame ref.) & 43.00 & 1.73 & 100.0 & \textbf{93.56} & \textbf{90.25} & \textbf{97.40} & 1.000 & 0.000 & -- & 0.0 \\
\midrule
Periodic dense reset ($K{=}4$) & \textbf{35.05} & 1.89 & 23.8 & 91.76 & 89.18 & 96.49 & 0.541 & 0.809 & 14.85 & 21.9 \\
Eventful reuse & 37.70 & 1.94 & 24.6 & 92.10 & 89.10 & 96.70 & 0.529 & 0.831 & 13.89 & 19.3 \\
Eventful reuse ($2\times$ budget) & 39.35 & 1.88 & 49.2 & 92.45 & 89.59 & 97.25 & 0.567 & 0.736 & \textbf{15.56} & 7.9 \\
\midrule
\textbf{Pre-RoPE recon. (matched)} & 36.37 & \textbf{1.76} & 24.6 & 92.77 & 89.85 & \textbf{97.40} & \textbf{0.552} & \textbf{0.780} & 14.92 & \textbf{8.9} \\
Pre-RoPE recon. (adaptive) & 36.63 & 1.80 & 23.8 & \textbf{92.79} & \textbf{89.87} & 97.25 & 0.552 & 0.779 & 15.01 & 11.5 \\
\midrule
+ Chained KV update (100\% ref.) & 38.92 & 1.71 & 100.0 & \textbf{93.64} & \textbf{90.90} & 97.10 & 0.586 & 0.700 & \textbf{15.84} & 5.9 \\
+ Chained KV + Eventful reuse & 34.69 & 1.85 & 24.6 & 92.54 & 89.91 & 96.63 & 0.551 & 0.768 & 14.55 & 9.1 \\
\textbf{+ Chained KV + Pre-RoPE recon.} & \textbf{32.85} & \textbf{1.66} & 24.6 & 93.41 & 90.68 & \textbf{97.39} & \textbf{0.597} & \textbf{0.688} & 15.77 & \textbf{6.3} \\
\bottomrule
\end{tabular}

\caption{\textbf{Complete 48-second results on the uniform-cache engine (12-frame cache; LongLive-1.3B, 8 prompts, 765 frames)}, expanding Tab.~\ref{tab:quality-transfer}a with adaptive Pre-RoPE and intermediate chained-KV rows.}
\label{tab:absquality-full}
\end{table*}

\begin{table*}[!t]
\centering
\small
\setlength{\tabcolsep}{2.9pt}
\begin{tabular}{@{}lrrrrrrrrr@{}}
\toprule
 & \multicolumn{3}{c}{Execution} & \multicolumn{3}{c}{No-Ref.\ Quality Diagnostics} & \multicolumn{3}{c}{Ref.-Trajectory Agreement} \\
\cmidrule(lr){2-4}\cmidrule(lr){5-7}\cmidrule(lr){8-10}
Method & DiT $\downarrow$ & Steps $\downarrow$ & Ref. & Diag. $\uparrow$ & Subj. $\uparrow$ & Smooth. $\uparrow$ & SSIM $\uparrow$ & AlexD. $\downarrow$ & PSNR $\uparrow$ \\
 & (s) & /tok & (\%) & (comp.) & (\%) & (\%) & & (norm.) & (dB) \\
\midrule
Full memory (reference) & 11.73 & 2.22 & 100.0 & \textbf{92.58} & \textbf{91.06} & 96.89 & 1.000 & 0.000 & -- \\
\midrule
No clean KV update ($0\%$) & \textbf{8.59} & 2.13 & 0.0 & 90.48 & 90.11 & 95.59 & 0.548 & 0.881 & 14.45 \\
Periodic dense reset ($K{=}4$) & 9.53 & 2.25 & 20.0 & 90.91 & 90.74 & 96.83 & 0.564 & 0.856 & 14.40 \\
Eventful reuse & 10.65 & 2.47 & 24.6 & 91.16 & 89.87 & 96.99 & 0.577 & 0.817 & 14.28 \\
Eventful reuse ($2\times$ budget) & 10.96 & 2.38 & 49.2 & 91.71 & 90.40 & 97.01 & \textbf{0.616} & \textbf{0.691} & \textbf{15.25} \\
\midrule
Pre-RoPE recon. (zero-shot \llcal{}) & 10.43 & 2.30 & 24.6 & 89.62 & 89.14 & 97.00 & 0.582 & 0.820 & 14.61 \\
\midrule
Pre-RoPE recon. (in-domain calib.) & 10.43 & 2.31 & 24.6 & 91.10 & 90.14 & \textbf{97.15} & 0.587 & 0.798 & 14.77 \\
Pre-RoPE recon. (in-domain + adaptive) & 10.70 & 2.35 & 32.1 & 90.91 & 90.08 & 97.11 & 0.592 & 0.782 & 14.87 \\
\midrule
+ Chained KV + adaptive (stacked) & 9.69 & 2.15 & 30.5 & 92.10 & \textbf{91.16} & \textbf{97.18} & \textbf{0.631} & \textbf{0.654} & \textbf{16.32} \\
\bottomrule
\end{tabular}

\caption{\textbf{Complete Self-Forcing-DMD-1.3B cross-backbone results (16 prompts, 189 frames)}, expanding Tab.~\ref{tab:quality-transfer}b with periodic reset and unchained adaptive rows.}
\label{tab:sfd-full}
\end{table*}

\begin{table*}[!t]
\centering
\small
\setlength{\tabcolsep}{1.9pt}
\begin{tabular}{@{}lrrrrrrrrr@{}}
\toprule
Variant ($24.6\%$ refresh unless noted) & DiT $\downarrow$ & Steps $\downarrow$ & PSNR$_{\text{all}}$ $\uparrow$ & $\Delta$PSNR & PSNR$_{\text{late}}$ $\uparrow$ & SSIM $\uparrow$ & AlexD. $\downarrow$ & Diag. $\uparrow$ & Sat. $\downarrow$ \\
 & (s) & /tok & (dB) & vs.\ ref. & (dB) & & (norm.) & (comp.) & (\%) \\
\midrule
Dense, block-1 noise $\epsilon{=}10^{-3}$ (100\% refresh) & 10.72 & 1.77 & 22.602 & -- & 20.38 & 0.810 & 0.289 & 94.01 & 4.2 \\
Dense, block-1 noise $\epsilon{=}10^{-2}$ (100\% refresh) & 10.69 & 1.78 & 21.911 & -- & 19.65 & 0.802 & 0.297 & 93.84 & 6.1 \\
\midrule
Eventful reuse (baseline) & \textbf{9.08} & 1.85 & 16.866 & -- & 15.57 & 0.625 & 0.608 & 93.29 & \textbf{7.2} \\
w/o age tracking + reuse-style selection & 9.76 & 1.96 & 16.575 & -0.463 & 15.14 & 0.630 & 0.631 & 92.48 & 11.4 \\
\midrule
Reference estimator (with context term) & 9.68 & 1.91 & 17.038 & 0.000 & 15.89 & 0.638 & 0.609 & 93.12 & 8.6 \\
\midrule
History only ($w \equiv 0$, matched selection) & 10.43 & 2.22 & 16.077 & -0.961 & 14.32 & 0.613 & 0.671 & 92.04 & 13.4 \\
In-flight only ($w \equiv 1$) & 9.90 & 1.96 & 16.386 & -0.652 & 15.53 & 0.622 & 0.631 & \textbf{93.70} & 11.9 \\
w/o Pre-RoPE affine transport ($W{=}I, b{=}0$) & 9.82 & 2.02 & 16.205 & -0.833 & 14.78 & 0.617 & 0.639 & 93.11 & 18.9 \\
Correlated BLUE gain ($(e_o{-}c_{or})/(e_o{+}e_r{-}2c_{or})$) & 9.96 & 2.02 & 16.295 & -0.743 & 15.06 & 0.622 & 0.660 & 92.72 & 13.2 \\
Unstratified transport (step-agnostic $W, b$) & 9.71 & 1.93 & 16.762 & -0.276 & 15.64 & 0.631 & 0.629 & 93.08 & 11.2 \\
w/o spherical norm restoration & 9.93 & 1.95 & 17.145 & +0.107 & 15.97 & \textbf{0.644} & 0.599 & 93.32 & 8.0 \\
\midrule
\textbf{Simplified Pre-RoPE recon.\ (w/o context term)} & 9.12 & \textbf{1.79} & \textbf{17.325} & \textbf{+0.287} & \textbf{16.29} & 0.643 & \textbf{0.580} & 93.45 & 8.7 \\
\bottomrule
\end{tabular}

\caption{\textbf{Complete estimator ablation on HeldOut-16 (189 frames)}, expanding Tab.~\ref{tab:est-ablation} with DiT wall time, $\text{PSNR}_{\text{late}}$, SSIM, and saturation drift. Row names follow Tab.~\ref{tab:est-ablation}; $\Delta$PSNR is the $\text{PSNR}_{\text{all}}$ difference to the reference estimator.}
\label{tab:est-ablation-full}
\end{table*}

\begin{figure}[!t]
\centering
\includegraphics[width=\linewidth]{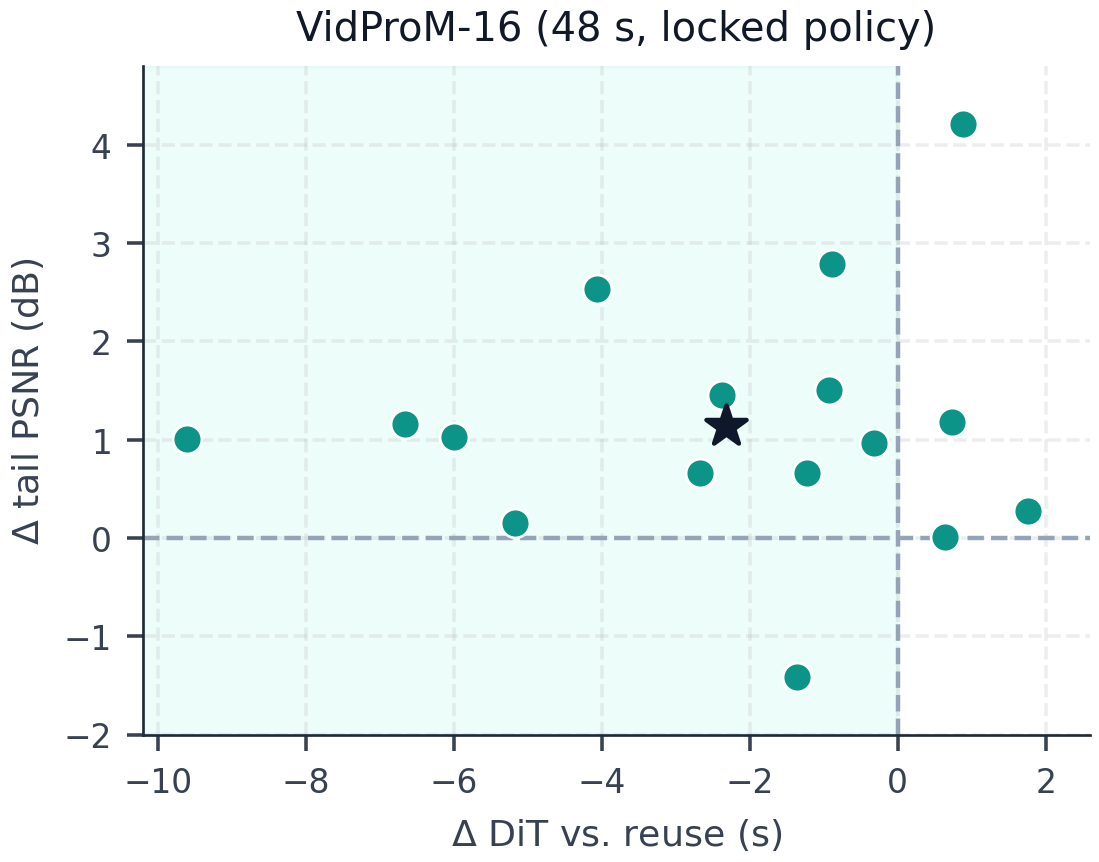}
\caption{\textbf{Per-prompt 48-second paired deltas on VidProM-16 (\method minus Eventful reuse, locked policy)}, complementing Tab.~\ref{tab:main}b (\UntouchedDitDelta{}\,s mean, $+\UntouchedGain{}$\,dB Q4 PSNR; \UntouchedPSNRWins{}/16 PSNR wins, \UntouchedSpeedWins{}/16 speed wins).}
\label{fig:untouched}
\end{figure}

\begin{table*}[!t]
\centering
\small
\setlength{\tabcolsep}{4.2pt}
\begin{tabular}{lrrrrrrr}
\toprule
Single-variable change & DiT $\downarrow$ & $\Delta$DiT & PSNR $\uparrow$ & $\Delta$PSNR & Flicker $\downarrow$ & $\Delta$Flk & Decision \\
 & (s) & (s) & (dB) & (dB) & (\%) & (pts) & \\
\midrule
\textbf{SAMR (locked policy)} & \textbf{30.11} & -- & \textbf{15.55} & -- & \textbf{18.01} & -- & Keep \\
Layer-balanced risk & 30.40 & +0.29 & 15.11 & -0.44 & 14.65 & -3.36 & Reject \\
Count hysteresis & 30.23 & +0.11 & 15.60 & +0.05 & 19.59 & +1.58 & Reject \\
Action downgrade & 30.08 & -0.04 & 15.56 & +0.01 & 20.35 & +2.34 & Reject \\
Step-1 reliability skip & 26.78 & -3.34 & 15.07 & -0.48 & 30.82 & +12.81 & Reject \\
Step-1 cooldown & 29.74 & -0.38 & 15.52 & -0.03 & 18.59 & +0.58 & Reject \\
Refresh-floor slope & 30.30 & +0.18 & 15.61 & +0.06 & 18.72 & +0.71 & Reject \\
\bottomrule
\end{tabular}

\caption{\textbf{Single-variable screen on Dev-8 (8 calibration prompts $\times$ 2 held-out seeds, 16 rollouts/arm)} against the locked \method policy (\ScreenDit{}\,s, \ScreenPSNR{}\,dB). Acceptance thresholds on DiT time, Q4 PSNR, and flicker drift were fixed before the runs; no candidate met them, so the policy was left unchanged. The latency weight $\lambda{=}2$ discussed in \S\ref{sec:ablation-text} is evaluated on \texttt{Main 15$\times$3} and is the last row of Tab.~\ref{tab:ablation}.}
\label{tab:screen}
\end{table*}

\paragraph{Contents and prompt splits.}
Tables~\ref{tab:config-map} and~\ref{tab:prompt-history} attribute each reported result to its engine configuration and prompt split. Tables~\ref{tab:exploratory}--\ref{tab:est-ablation-full} give the exploratory variants and the unabridged versions of the main-paper tables, Fig.~\ref{fig:untouched} the per-prompt \texttt{VidProM-16} deltas, and Tab.~\ref{tab:screen} the \emph{Dev-8} screen. \texttt{Main 15$\times$3} draws its prompts from the 16-prompt held-out list; one prompt was excluded before any run because of its political content, leaving $15$ prompts in every 45-rollout experiment. HeldOut-16 denotes the full 16-prompt list at 12\,s (189 frames, 16 blocks), and Long-64 denotes 64-block rollouts (765 frames, 48\,s) of eight of its prompts.

\paragraph{Equal-latency comparison.}
For the equal-latency comparison in \S\ref{sec:absolute}, the refresh count of \emph{Eventful reuse} is chosen from $m \in \{768, 896, \dots, 1408\}$ on \emph{Dev-8} only (seed $730001$, two repeats), using latency alone: we take the $m$ that minimizes $|\log(T_{\mathrm{reuse}}/T_{\mathrm{assim}})|$ against fixed-budget \emph{Calibrated assimilation}. This selects $m=1{,}152$, \ie the default $24.6\%$ budget. On \texttt{Main 15$\times$3} the two then run within the $5\%$ tolerance ($30.15$ \vs $29.46$\,s), and \emph{Calibrated assimilation} retains $+1.05$\,dB Q4 PSNR ($95\%$ prompt-bootstrap interval $[+0.69, +1.45]$\,dB).

\section{Additional Qualitative Comparisons}
\label{sec:qualitative-extra}

\begin{figure*}[!t]
\centering
\includegraphics[width=0.95\textwidth]{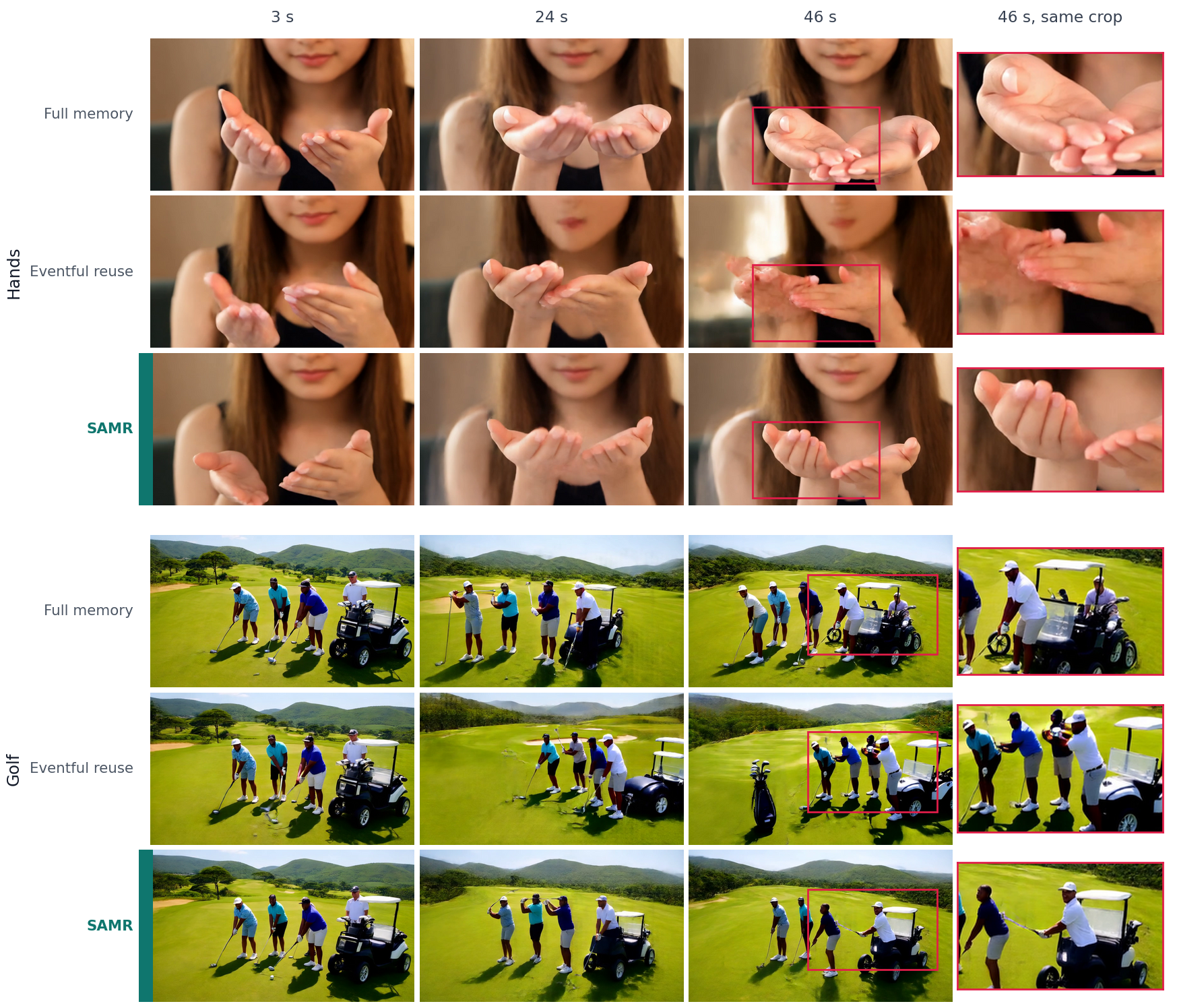}
\caption{\textbf{Complete 4-timestamp qualitative strips for \emph{Hands} and \emph{Golf} (seed 0; matched $3$\,s, $24$\,s, $46$\,s, and $46$\,s crop)}, expanding Fig.~\ref{fig:qualitative} with intermediate $24$\,s frames.}
\label{fig:qualitative-full}
\end{figure*}

\begin{figure*}[!t]
\centering
\includegraphics[width=0.95\textwidth]{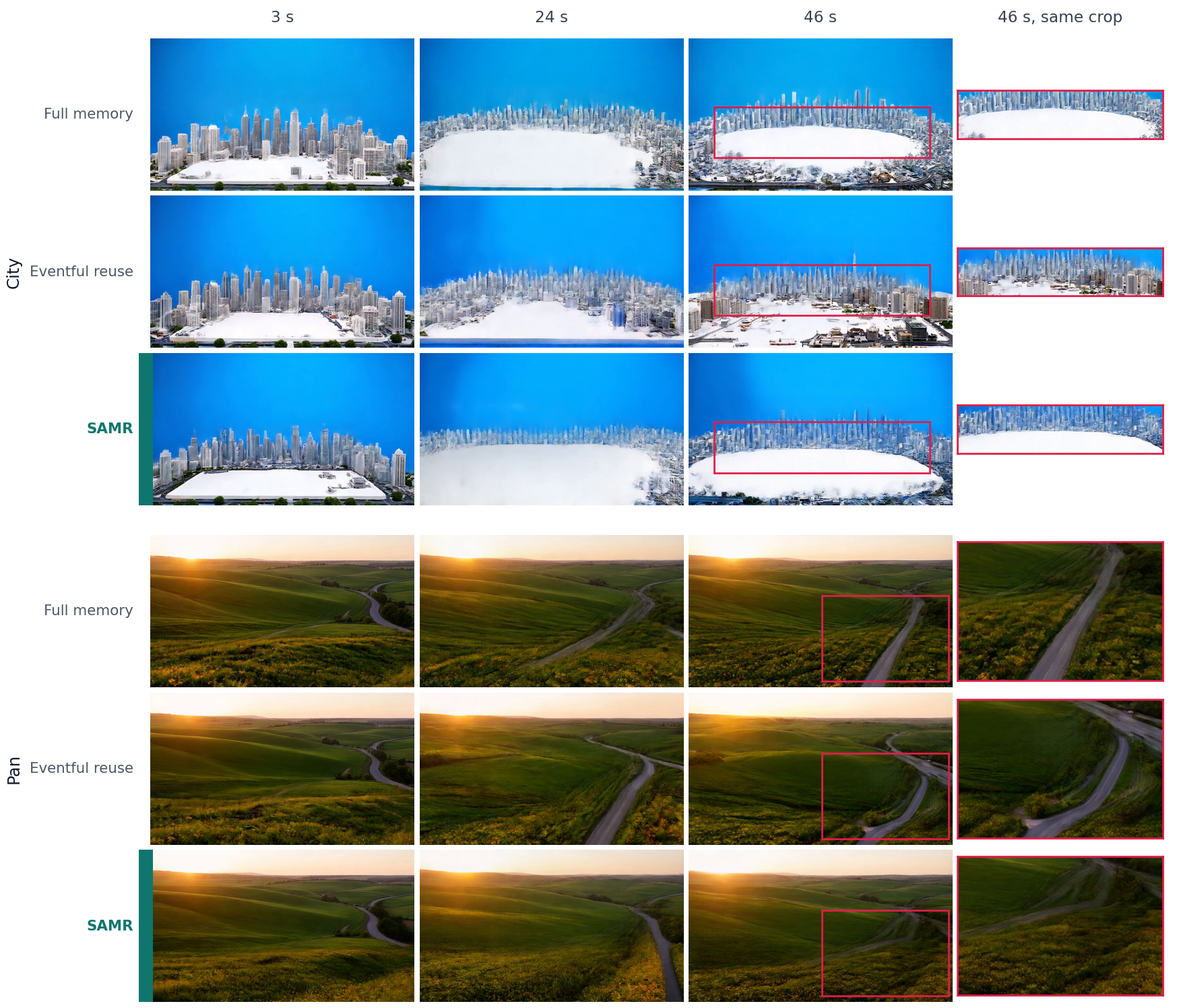}
\caption{\textbf{Additional same-seed 48-second qualitative rollouts under non-stationary evolution and sustained camera motion (seed 0; matched $3$\,s, $24$\,s, and $46$\,s timestamps).} On \emph{City} (growing city timelapse), naive reuse fragments skyscraper geometry, whereas \method preserves architectural boundaries while lowering DiT time from reuse's $24.13$\,s to $22.43$\,s ($1.31\times$ faster than \emph{Full memory}'s $29.29$\,s). On \emph{Pan} (countryside camera pan), copying prior KV leaves a ghosted double-highway ($20.23$\,dB Q4 PSNR, $24.82$\,s), whereas \method elevates refreshes along newly entering road borders ($31.3\%$ mean refresh) and reconstructs rolling hills ($22.19$\,dB, $+1.96$\,dB over reuse, $22.85$\,s). Times and PSNR are three-seed means for the prompt.}
\label{fig:qualitative-extra}
\end{figure*}

Figures~\ref{fig:qualitative-full} and~\ref{fig:qualitative-extra} provide the un-cropped 4-timestamp progression ($3$\,s, $24$\,s, $46$\,s, and $46$\,s zoom) for \emph{Hands} and \emph{Golf} alongside two additional 765-frame seed-0 rollouts stressing continuous camera panning (\emph{Pan}) and structural synthesis (\emph{City}). All frames are extracted losslessly from pre-encoding RGB evaluation outputs without post-processing outside the indicated zoom box.

\section{Extended Discussion of Limitations}
\label{sec:ext-limitations}
\paragraph{Risk model.}
The readout proxy (Eq.~\eqref{eq:readout}) is an offline, single-token, first-order estimate: it does not observe live queries and does not bound multi-block error. The staleness recurrence (Eq.~\eqref{eq:lineage}), the residual refresh floor, and the latency model are empirical rather than derived.

\paragraph{Cross-backbone transfer.}
On Self-Forcing-DMD-1.3B, fixed-budget reconstruction with LongLive maps transfers zero-shot in steps and PSNR, but \emph{Diag.} drops below reuse ($\SFDReuseVB{} \to \SFDZeroshotVB{}$). A one-time, training-free in-domain calibration recovers it to $\SFDIndomainVB{}$, which is still below dense updates ($\SFDFullVB{}$); stacking with chained KV updates and adaptive budgeting reaches $\SFDCFDynVB{}$ (Tab.~\ref{tab:sfd-full}). We have not transferred the full adaptive LongLive policy zero-shot.

\paragraph{Statistical resolution.}
On \texttt{Main 15$\times$3}, the paired intervals for DiT time and Q4 PSNR against reuse exclude zero, but those for Q4 saturation, flicker, and motion drift span it (Tab.~\ref{tab:paired}), and fixed-budget \emph{Calibrated assimilation} attains lower mean drift than full \method. \texttt{VidProM-16} uses a fresh seed, but its prompt texts come from the \mbox{12-s} \texttt{VidProM-64} pool (Tab.~\ref{tab:prompt-history}).

\paragraph{Evaluation scope.}
Our kernels target 1.3B DiTs on AMD GPUs. We report reference-trajectory agreement and reference-free diagnostics; standard LPIPS~\cite{zhang2018lpips}, FVD, official VBench-Long~\cite{huang2024vbench}, user studies, and direct runtime comparisons with FlashForward~\cite{wang2026flashforward} and ARCache~\cite{nan2026arcache} remain future work.

\end{document}